\documentclass[graybox]{svmult}
\usepackage{enumerate}
\usepackage{tikz}
\usepackage[sort, numbers]{natbib}
\usepackage[ruled]{algorithm2e}
\usepackage{xcolor}
\usepackage{rotating}
\usepackage{graphicx}
\usepackage{epsfig}	
\usepackage{subfigure}	
\usepackage{wrapfig}
\usepackage{pdflscape}
\usepackage{tabularx}
\usepackage{color, colortbl}
\usepackage{float}
\restylefloat{table}
\usepackage{esvect}   
\usepackage{soul}
\usepackage{algpseudocode}  
\usepackage[bottom]{footmisc}
\usepackage{multicol}
\usepackage{multirow}
\usepackage{diagbox, eqparbox, hhline}

\usepackage{amsfonts}
\newcolumntype{L}[1]{>{\raggedright\let\newline\\\arraybackslash\hspace{0pt}}m{#1}}
\newcolumntype{C}[1]{>{\centering\let\newline\\\arraybackslash\hspace{0pt}}m{#1}}
\newcolumntype{R}[1]{>{\raggedleft\let\newline\\\arraybackslash\hspace{0pt}}m{#1}}
\definecolor{Gray}{gray}{0.9}

\begin{document}
\title{Visual Methods for Sign Language Recognition: A Modality-Based Review}
\author{Bassem Seddik  \and Najoua Essoukri Ben Amara }
\institute{Bassem Seddik  \and 	Najoua Essoukri Ben Amara 	
	%\\
	\at Universit{\'e} de Sousse, Ecole Nationale d'Ing{\'e}nieurs de Sousse, LATIS- Laboratory of Advanced
	Technology and Intelligent Systems, 4023, Sousse, Tunisie;
	\\
	\email{bassem.seddik@vizmerald.com, najoua.benamara@eniso.rnu.tn}
	}

\maketitle
\vspace{-0.7cm}
\abstract
{
 Sign language visual recognition from continuous multi-modal streams is still one of the most challenging fields. 
 Recent advances in human actions recognition are exploiting the ascension of GPU-based learning from massive data, and are getting closer to human-like performances. 
 They are then prone to creating interactive services for the deaf and hearing-impaired communities.
 A population that is expected to grow considerably in the years to come. 
 This paper aims at reviewing the human actions recognition literature with the sign-language visual understanding as a scope. 
 The methods analyzed will be mainly organized according to the different types of unimodal inputs exploited, their relative multi-modal combinations and pipeline steps. 
 In each section, we will detail and compare the related datasets, approaches then distinguish the still open contribution paths suitable for the creation of sign language related services.
 Special attention will be paid to the approaches and commercial solutions handling facial expressions and continuous signing. 
}

\section{Introduction}

	Sign language (SL) is a purely visual communication manner relying on a combination of hand gestures, facial expressions and upper-body postures. For nearly 466 million persons world-wide, SL is their first native language. This is particularly the case for those born with total or partial hearing loss, thus with limited or no speech capabilities. 	
	It is a considerable population that suffers from isolation, inadequate education and very few life services. 
	
	This is specially emphasized as most hearing/speaking persons are not able to interact with SL natives. Despite that devices and implants were invented for audition increase, still they are very far from allowing a good hearing. Thus interaction through SL remains a vital need since the youngest babies age \cite{heringLossStats2013}. 
	According to the world health organization, the lack of deaf/hearing-impaired services and employment is causing a global annual loss of 75 billion USD \cite{WorldHealthOrganization}. This is an increasingly crucial situation as nearly 900 million persons are expected to show hearing loss by 2050. This explains the importance of considering SL recognition from a technically visual point of view. A situation purely within the computer vision field specialization. 
		
	 Here are the key-word terminologies that we apprehend in our review: 
	 We distinguish different approach families related to the task of human actions recognition. 
	 Each one of these is concertized using a specific methodology composed itself of a sequence of consecutive steps leading to the desired result.
	 Within a finer analysis scale, we find that each step could be implemented using a set of {techniques} according to selected programming {algorithms}.
	 In the case where all the {steps} of a {method} belonging to a given {approach}, are implemented from end to end, we consider that we have a fully operational  {system} of some originality.
	 We will also refer with a capital ``D'' to the Deaf community as most deaf persons strongly identify to this culture.

		\begin{figure}[!t]
			\centering
			\subfigure[]{	
				\includegraphics[height = 0.175\textwidth]{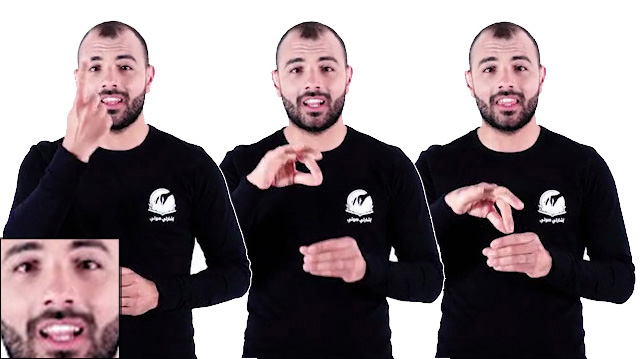}
			}
			\subfigure[]{	
				\includegraphics[height = 0.175\textwidth]{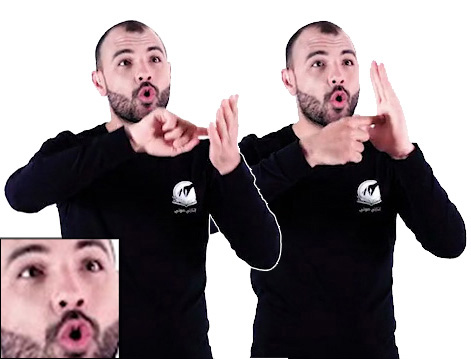}
			}	
			\subfigure[]{	
				\includegraphics[height = 0.175\textwidth]{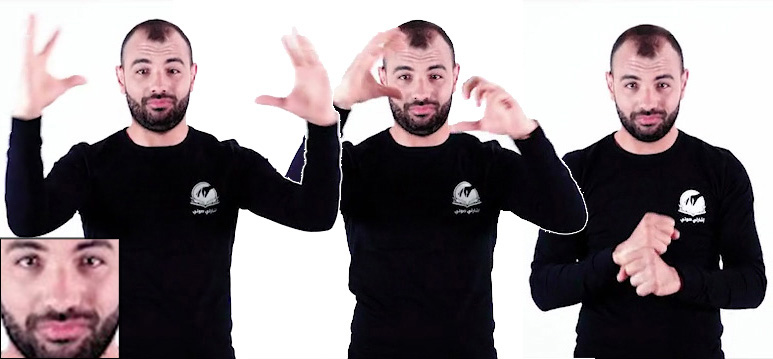}
			}	
			\caption{\small Importance of the face and hand gestures to the sign-language lexicon: (a) One-vote, (b) verify, (c) collect-all \cite{IcharatiSawti2019}}
			\label{fig:LSecrite}
		\end{figure}

	 		The next coming sections are logically organized but also self contained for an easier reading. 
	 		%for separately consult those of interest 
	 		In section 2, we will start by reviewing the general pipeline adopted for human actions recognition. 
	 		In section 3, we will analyze the existing solutions that rely on unimodal inputs. Section 4 will then be dedicated to multi-modal approaches.
	 		%and multi-modal those exploiting existing joint positions with RGB 
	 		For each of the mentioned sections, we will be pointing the main contributions suitable for SL recognition. An analysis of the commercial solutions actually targeting the Deaf community will be presented in section 5. We will rely on it for the proposal of key axis of contribution that could be developed by future techniques. 
	\\
	 During our state of the art analysis, we will additionally focus on facial expressions and hand gestures recognition as they strongly contribute to the definition of various SL lexicons \cite{SignLangFramework}. As shown in Fig.\ref{fig:LSecrite}, hand and face movements are sometimes decisive for distinguishing actions.

\section{Human actions recognition pipeline}
%\vspace{-0.1cm}
 {Literature works on SL recognition can be classified according to the different stages of their processing pipeline. While the early approaches were based on static captures \cite{SurveyRGB2014, Fasel_and_Luettin_2003}, the actual trend is towards the recognition from continuous multi-modal streams and large learning populations \cite{Survey3D2014, Survey_FacialExp_2012}. Thus, several additional processing steps became influential to Human Actions Recognition (HAR) in general and SL in particular.}

 {The existing approaches' pipelines could generally be classified according to their: (i) input data, (ii) pre-processing steps, (iii) features types extracted, (iv) data representations, (v) temporal segmentation techniques, (vi) classification methods as well as (vii) post-processing outputs producing the final labels \cite{SurveyHAR_2015}. We remind hereafter the main details of these commonly used steps:}
 
 % généralement les étapes les plus suives

\begin{enumerate}[i]
	\item \textbf{Input data:} They determine the mass and type of the raw input streams within a given method. These streams could be composed of joints, RGB, depth or mixtures of these modalities. 
	We can also opt for a vocabulary related, for example to sign language, daily actions or facial expressions.
	\item \textbf{Preprocessing:} This step implies the usage of any filtering, normalization and data formatting operations in order to improve its readability. Due to multi-modal sources usage, the streams synchronization became also a very important step to ensure their correct subsequent processing.
	\item \textbf{Temporal segmentation:} If we consider recognition from continuous action streams, it is recommended to first segment them into partial sub-actions. This is done by searching the pairs of (beginning, 	end) temporal positions for each significant sub-gesture relative to the vocabulary. %Once the sub-gestures separated, recognition is affordable later.	
	\item \textbf{Feature extraction:} This step applies calculations, measures and vectors extraction to form a more consistent meta-information compared to the raw input data. These descriptors can be generated by sophisticated extractors obeying to specific spatio-temporal sampling steps, or more recently, pooled from deep neural-network models \cite{DeepSurvey2016}.
	\item \textbf{Representation:} In order to further improve the homogeneity and efficiency of the obtained descriptors, additional steps of data representation could be added. They are particularly valuable in the case of learning from large scale populations (e.g. +10K sequences \cite{CGC2011Article}).
	\item \textbf{Classification:} In this step, the extracted information is separated into groups belonging to the different action labels.
	\item \textbf{Post-processing:} In an even more advanced step, there are additional interesting post-processing tools for merging and optimizing the decisions obtained from classifiers through different fusion strategies.
\end{enumerate}

 {In this paper, our literature classification will be based on the modalities data-types used. We will distinguish between unimodal vs multi-modal approaches and review their main contributions at each of the aforementioned steps. For each data input, we will especially focus on the respective description, segmentation and post-processing stages techniques applied.}

\section{Unimodal methods}

As the Kinect sensor has been used extensively for acquisition in recent works, synchronized joint, {RGB} and {depth} streams became simultaneously available. Despite this, numerous contributions focused solely on a unique modality among them. 
We begin this unimodal-dedicated section by presenting the joint-based works. We will then analyze uni-modal literature dedicated to RGB streams and finish with those based on depth. We highlight that we consider ``unimodal'' the approaches adding hand-made annotation points to RGB available data.

%italic!!!
\subsection{Recognition from joint streams}	

{The first joint-based works were based on motion sensors (i.e. Motion Capture (MoCap)) of different types attached to the actor's body. In this configuration, the person (i.e. the signer) is the egocentric carrier of the capture device \cite{EgoCentricVideos2014}. 
Another family of MoCap technologies is based on a remote network of sensors. The use of landmarks (i.e. {marker-points}) is then generally affordable using infra-red capture devices such as in \cite{CMU_manifold2010}. A review of MoCap-based techniques for HAR is presented in \cite{SurveyWearbleSensors2013}.}

\begin{figure}[t]
	\centering
	\subfigure[OpenNI]{
		\includegraphics[height = 0.25\textwidth]{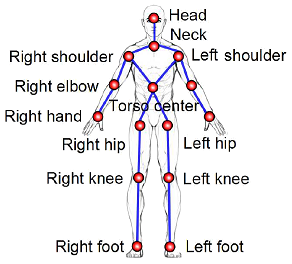}	
	}
	\subfigure[Kinect-V1]{
		\includegraphics[height = 0.25\textwidth]{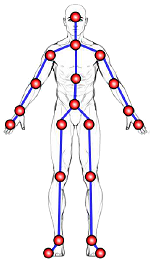}	
	}
	\quad
	\subfigure[MoCap]{
		\includegraphics[height = 0.25\textwidth]{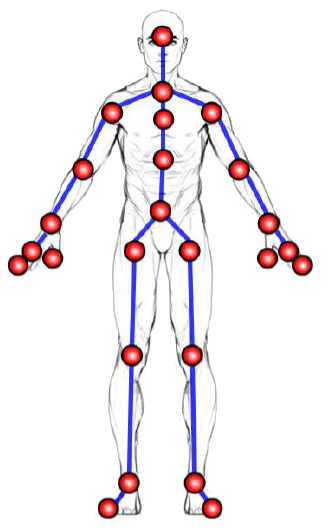}	
	}		
	\caption{\small Examples of joint 3D body-landmarks increasing by technology \cite{jointSolutions_2016} }
	\label{fig:SkeletalRepresenations}
\end{figure}

 {With the apparition of infra-red sensors similar to the Kinect, marker-less captures become easily achievable by the computer vision community \cite{jointSolutions_2016}.
As a result, various configurations of joints have been proposed based on 15, 20 and 25 positions (among others, see Fig. \ref{fig:SkeletalRepresenations}) with technologies such as OpenNI, Kinect and MoCap, respectively.
A large number of contributions have relied solely on joints to provide very accurate performances. An explanation was provided by Jhuang's {et al.} study \cite{Jhmdb_Act_Recog_Undesrstand2013}: The articulation as a high order modality allows (in general) gains of 19 to 29\% of accuracies compared to mono-modal configurations in {RGB} or {depth}.
}

\subsubsection{Joint-based datasets}  		
 {One the most used datasets is the MSR-Action3D \cite {ActionLet2012}. It has been first introduced with a depth-based approach, but its joint streams were widely used later as a reference for joint-based methods. And while different training schemes have been adopted, the most difficult scenarios are those applying Cross-Validation (CV) evaluations.}

% les complexités des bases avec des images, voici les cas à challenges
% indicaiton sur les perfomances et les classifieurs
%

 {A similar interesting CV evaluation is also found within the CAD-60 database \cite{CAD60_Sung2012}. It offers daily-action human movements categorized by rooms. A similar multi-modal dataset is MSR-Daily. A significant information is related to the streams durations. While CAD-60 has a small number of samples, it has the merit of offering streams that span over relatively long periods of time: 50 to 120 seconds. The larger is the frame samples number, the higher is the likeliness level for intra-class variability. This is also the case within large-scale datasets offering joint streams such as those relative to the Chalearn Gesture Challenges (CGC) \cite{CGC2014trac3}. At the moment of this survey writing, one of the largest benchmarks offering body joints is the NTU-RGB+D \cite{Kinect2_largeScale2016}. Though this dataset offers many other modalities, recent methods are using it to evaluate their joint-only-based performances.
 	
  Estimating joint positions from visual data has also contributed to the creation of specialized datasets such as FLIC \cite{Modec2013} or MPII \cite{MPIIdataset2014}. Lately, human pose estimators (i.e. position, rotation and scale in 2D or 3D spaces) such as the OpenPose \cite{OpenPose_2018} have received interest for the generation of joint data from large-scale RGB datasets. This new trend of estimating pose landmarks from in-the-wild images is prone to further developments \cite{PoseEstimation_Shah2015, Pose_Zisserman2016, DepthPose_FeiFei2016, densePose2018} as the pose-estimated joint performances are still lower than those sensor-based. This is for instance the case for the state-of-the art approach of Shi et al. \cite{DGNN_Joint2019} with only 36.9\% as recognition rate when using the skeletons generated from the Kinetics RGB dataset \cite{Kinetics2017_dataset}. With the development of synthetic 3D datasets dedicated to human pose estimation such as SURREAL and DensPose \cite{surrealDataset2017, densePose2018} promising paths for SL recognition in-the-wild are still expected. A detailed analysis of pose-based HAR is provided by Boualia et al. in \cite{SamehNeili2019}.
  For what follows in this review, we only consider this category of works using pose estimation for the main purpose of actions recognition. 

\subsubsection{Joint-based approaches for HAR}					
 {The first approaches dedicated for human movements description were derived from the {Shape-Context} (SC) descriptor initially proposed by \cite{BelongieMP02} to describe 2D forms in quantized discrete points. The SC made it possible to generate histograms relative to the distances and angles between the various considered points. A successful extension of the SC descriptor to generate histograms of 3D joints has been proposed in \cite{SC_Joints2012}. Figure \ref{fig:ShapeContext} illustrates examples of SC-based approaches in 2D and 3D. }

 {In the representation based on principal joints ({EigenJoints}), the authors based their contribution on cumulative energies descriptor from joint movements \cite{EigenJoint_2014}. They applied Principal Component Analysis (PCA) representation and a nearest-neighbor classification.
A very close approach has been proposed in \cite{SMIJ_joints_2014}. The authors start by extracting the most discriminative articulations to concatenate their cumulative motion histograms. A related enhancement is proposed in the descriptor {ActionLet} \cite{ActionLet2012}. In addition to the movements cumulation and histograms concatenation, the authors have also extracted the temporal pyramid Fourier coefficients to manage the actions' misalignments.}

 {A richer set of cumulative features have been proposed in the {MovingPose} \cite{MovingPose2013} representation. The authors propose to compute first-order and second-order motion gradients (i.e. velocities and accelerations) then concatenate N groups of frame contents into unique descriptors per key-frame. The set of descriptors is combined with an SVM classifier to give 91.7\% performances on the MSR-Action3D database within its most challenging cross-validation configuration.}
 
\begin{figure}[!t]
	\centering
	\subfigure[\ ] {\includegraphics[width = 0.26\textwidth]{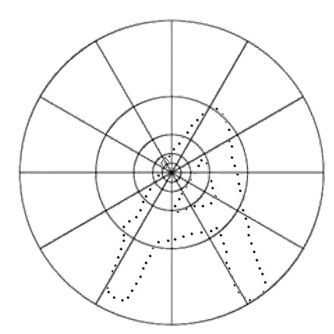}
		%\label{fig:SC_2D}	
	}\qquad
	\subfigure[\ ] {\includegraphics[width=0.26\textwidth]{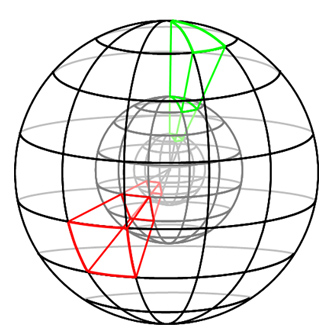}
		%\label{fig:SC_3D_Sphère}	
	}\qquad
	\subfigure[\ ]
	{\includegraphics[width=0.3\textwidth]{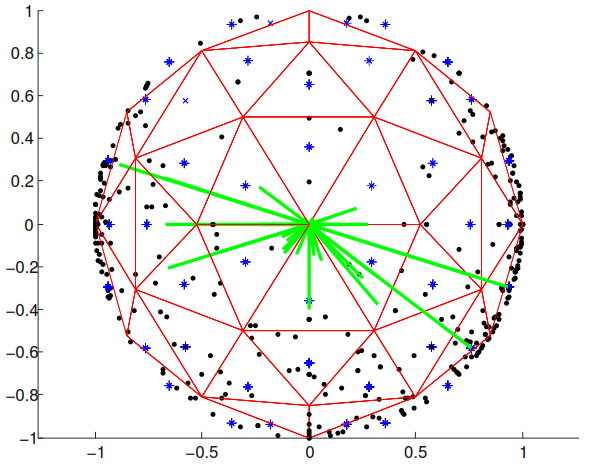} }%\label{fig:hog3D}			
	%\qquad
	%\subfigure[Contextes de formes 3D cylindriques \cite{Gond_these2009}] {\includegraphics[width=0.26\textwidth]{./figures/SC3DCylindre.jpg}
	%\label{fig:SC_3D_Cylindre}	
	%}
	\caption{\small Examples of SC-based descriptors: (a) in 2D \cite{BelongieMP02}, (b and c) in 3D contexts \cite{3DShapeContext_2010, klaser3DHOG2008} }
	\label{fig:ShapeContext}
\end{figure}

 {The best joint-only-based performance under the CGC-2014 dataset \cite{CGC2014trac3} is provided by Evangelidis {et al.} \cite{jntQuadruplets}. They used a selective concatenation of skeletal {Quads}: Each set of 4 joints is encoded and normalized according to its initial join. Thus, these {Quads} are designed to analyze the separate movement of arms, feet, torso, etc. In addition, the authors have applied the Fisher-Vectors (FV) representation to gain adaptability to large-scaled data.}

 {More advanced approaches add extra layers of specially designed data representations. For example, specific representation spaces referred as {Manifolds} hold the best performances reported on the CMU \cite{CMU_manifold2010} dataset. Methods such as the  {Kendall's Manifold} \cite{BenAmor_TPAMI2016} or the {lie-groups} \cite{Vemulapalli2016} have also offered excellent performances. Most of them define specific decision measures in the form of distances adapted to the considered spaces.}

 {In Table \ref{tab:Joint_Datasets_Performances}, we present a summary of the progressive contributions proposed to recognize human actions from join-only streams. Following a chronologically ascending order in Tab.\ref{tab:Joint_Datasets_Performances}, we note that the most recent approaches have gradually exploited most of the previous contributions. This is the case for the two most recent approaches presented in Tab.\ref{tab:Joint_Datasets_Performances}. For instance, the AGC-LSTM approach uses an efficient type of Recurrent Neural Network (RNN) known as LSTM (i.e. Long Short-Term Memory)  to enhance the convolutional graph while encoding the temporal actions evolution just as in the classic MovingPose descriptors \cite{MovingPose2013}. The top performance is reported on the NTU-RGB+D dataset by the Directed Graphs \cite{DGNN_Joint2019}. They achieve this as they encapsulate the joint kinetic relations within their deep neural network. From this trend, it is clear that the next performance improvements will be guided by how much of the classic hand-crafted features are simulated using the new deep neural architectures.}

  	\begin{table}[!t]
  		\caption{\small A selection of joint-based approaches for HAR and their relative datasets \cite{SurveySkeleton2015, Survey3D2014, SurveyWearbleSensors2013}. 1/2, 2/3 and CV: Use of one, two-thirds and cross-validation for learning, respectively}
  		\centering
  		\scalebox{0.8}{
  			\begin{tabular}{|c|C{2cm}|L{5.2cm}|c|C{4cm}|}
  				\hline
  				\textbf{\textsc{Year}} &            \textbf{\textsc{Approach/ Ref.}} & {\textbf{\textsc{Main contribution }}}                                           & \textbf{\textsc{Classifier}} &  {\textbf{\textsc{Performance per dataset}}}  \\ \hline\hline
  				2019          &               {DGNN} \cite{DGNN_Joint2019} & 2-stream convolutional network that encodes the joint kinematics                 &         Directed CNN         & 96.1\% NTU-RGB+D CV, 36.9\% Skeleton-Kinetics \\ \hline
  				2019          &         {AGC-LSTM} \cite{AGC_LSTM_Joint19} & Description of the spatial an temporal co-occurrence using LSTM networks               &           LSTM CNN           &              95.0\% NTU-RGB+D CV              \\ \hline
  				2016          &          {LieCurve} \cite{Vemulapalli2016} & Actions representation as points on a lie-curve and rotation quaternions         &             SVM              &  93.5\% MSR-Action3D CV,  97.1\% UTK-Action   \\ \hline
  				2014          &        {Quadruplets} \cite{jntQuadruplets} & Consideration of 4-joint groups and FV representation                            &             SVM              &                76.8\% CGC-2014                \\ \hline
  				2014          & Shan and Akella \cite{CAD60_Shan_ARSO2014} & Duplication by symmetric joints to improve performance                           &            RF/SVM            &      84\% MSR-Action3D CV, 93.8\% CAD-60      \\ \hline
  				2014          &       Ofli et al.  \cite{SMIJ_joints_2014} & Selection of the most informative joints from {EigenJoints}                      &      $\chi^2$ distance       &             79.9\% Berkeley-MHAD              \\ \hline
  				2014          &        {EigenJoint} \cite{EigenJoint_2014} & Cumulative movement energies over time then application of PCA                   &              BN              &            97.8\% MSR-Action3D 2/3            \\ \hline
  				2013          &         {MovingPose} \cite{MovingPose2013} & Normalization joint dimensions and concatenate $1^{st}$ and $2^{nd}$ order gradients &             SVM              &            91.7\% MSR-Action3D  CV            \\ \hline
  				2012          &                {H3DJ} \cite{SC_Joints2012} & Description by histograms of distances and orientations ({Shape Context})        &             HMM              &            96.2\% MSR-Action3D 1/3            \\ \hline
  				2012          &           {ActionLet} \cite{ActionLet2012} & Analysis in Fourier pyramids to temporally partition the action                  &             SVM              &           85.6\% MSR-Daily-Activity           \\ \hline
  				2010          &         {Manifold} \cite{CMU_manifold2010} & Transformation of inertial data to a surface-based space ({manifold})            &           RF + SVM           &               98.3\% CMU-mocap                \\ \hline
  			\end{tabular}
  		}
  		\label{tab:Joint_Datasets_Performances}
  	\end{table}

%\vspace{-0.1cm}
%\subsection{Travaux dédiés aux actions humaines}
\subsection{Recognition from RGB streams}

 {When analyzing the approaches based on RGB visual data (images or videos), it is noticeable that this modality is the most developed one due to its accessibility. Most contributions related to human actions from RGB streams focused mainly on feature extraction. While early contributions considered hand-designed features, the recent ones are relying more on variants of the Convolutional Neural Networks (CNN) architecture implemented successfully for the first time by Krizhevsky et al. in \cite{ImageNet2012} to produce automatically learned descriptors from massive data. We present in the following the RGB most used datasets and their related approaches evolution, both for actions and expressions recognition.}

\subsubsection{RGB-based datasets for HAR}

	The first RBG-based human action datasets presented in Tab.\ref{tab:DatsetRGB} focused on using fixed images acquired in the wild. Partially promoted by the VOC challenges \cite{Everingham10}, the best recognition rates for this task started at $45.7$\% for the {Stanford-40-Actions} dataset \cite{StanfordDataset_2011} to reach $74.2$\% for the {89-Action-Dataset} \cite{89Action_Dataset}.  	
	Soon after, a number of video-based datasets started offering labeled action sequences. 
	The {KTH} \cite {KTHdataset_2004}, {Weitzmann} \cite{WeizmannDataset2007} and {CAVIAR} \cite{SurveyRGB2013} datasets have been in used since {the 90's}. With over a hundred references each \cite{SurveyRGB2013}, they allow performances close to 100\% with nowadays methods.
	
	Known the relative facility of these first video benchmarks \cite{SurveyHAR_2015}, the community's concentration shifted towards video sequences with more ``confusing'' (i.e. real-word) content. We find for instance the {Olympic} dataset \cite{Olympic_2010}, the iterative improvements of {UCF} \cite {UCF50_2013} and Hollywood benchmarks \cite{Hollywood2Dataset_2009} in addition to the {HMDB} dataset \cite{ HMDB_dataset}.  
	Several other datasets such as i3DPost \cite{SurveyRGB2013}, MuHaVi \cite{MuHaVi2010} and HumanEva \cite{HumanEva_dataset} focused on multi-view and richer visual acquisition contexts (i.e. video streams resolutions,  silhouettes masks generation, etc.). In this case, the older contributions reached more than 80\% performance, while the most challenging MuHaVi dataset allowed $58.3$\% accuracy with the RGB-only approach presented in \cite{MuHaVi2010}.
	
	%Other data enrichments concerned also the labels and samples numbers (e.g. , {CAD}\cite {CAD60_KoppulaGS13} and {CGC}\cite{ConGD_IsoGD_cvpr2016} benchmarks).
	
	On another level, the time-line evolution clearly shows the trend for increasing the learning masses and the action labels: (i)
	On one hand, after having first used datasets offering hundreds of sample actions \cite{KTHdataset_2004}, then to the order of 1K learning actions such as UCF \cite{UCF50_2013}, the 2011--2014 contributions have been dealing with more than 10K videos samples as in CGC \cite{CGC2014trac3}. 
	(ii) On another hand, the number of action labels has gradually increased from 203 in the {ActivityNet} dataset \cite{ActivityNet_dataset2017} to 399 in the Moments-in-Time one \cite{MomentsTimedataset2018} and more recently is reaching 700 within the latest update of the Kinetics dataset \cite{Kinetcs700dataset_2019}.
	According to Carreira et al. in \cite{Kinetcs700dataset_2019}, it is expected to scale up 1K labels and 1M sequences (or to very long videos) in coming years. This is the case for instance with the Sports-1M \cite{MomentsTimedataset2018} and the ActivityNet \cite{ ActivityNet_dataset2017} datasets, respectively, enabling the CNN architectures learning. 
	
	For all of these recent benchmarks, the most advanced methods are limited to accuracies between $44.7$\% and $64.8$\%. To overcome this limitation, actual evaluation metrics often consider the top-5 correct recognitions. Another promising path is learning the actions from weakly-supervised sequences \cite{2020Trend, EPICKITCHENS2018}. Interestingly, this allows HAR using much larger amounts of noisy social-media sequences and hashtag labels (e.g. 65M sequences and 10.5K labels as in \cite{2020Trend}). In this case, the labels are rather combinations of verb-noun couples as employed by the EPIC-Kitchens dataset \cite{EPICKITCHENS2018}. This field in particular remains wide open to contributions, specially in first-person related streams \cite{SurveyFirstPerson2015}.

	%As shown in Tab.\ref{tab:DatsetRGB}, 
	In Table \ref{tab:DatsetRGB}, we consider the chronological performances evolution of different RGB-only approaches applied to public fully-labeled benchmarks. Although superior performances exist in the literature, we only maintained those in-line with the evaluation process of the dataset creators \cite{HMDB_dataset}. 
	
	The literature development of these datasets allowing RGB-only approaches is continued within multi-modal acquisition contexts under the section \ref{sec:mulmodalDatasets} and Tab.\ref{MultimodalWorks}.

\begin{table}[!t]
	\caption{\small Main datasets allowing RGB-based approaches for HAR \cite{SurveyFirstPerson2015, SurveyHAR_2015, SurveyRGB2014, SurveyRGB2013}. ($ * $) Indicates an adaptation of the measure used as a percentage.}
	\centering
	\scalebox{0.8}{
		\begin{tabular}{|C{1cm}|c|c|c|l|c|}
			\hline
			\textbf{ \textsc{Year}} & {\textbf{\textsc{Dataset}}} &   {\textbf{\textsc{Ref. }}}    & \textbf{\textsc{Performance (classifier)}} & {\textbf{\textsc{Contents}}}   & {\textbf{\textsc{Nb. labels}}} \\ \hline\hline
			                                                                            \multicolumn{6}{c}{\small From video streams}                                                                             \\ \hline
			        {2019}          &        Kinetics-700         & \cite{Kinetcs700dataset_2019}  &            58.7\% (LSTM + CNN)             & 600 seq. of 10s. by action     &              700 actions              \\
			        {2018}          &       Moments in Time       & \cite{MomentsTimedataset2018}  &       44.7\%   (spatiotemporal CNN)        & 1M seq. of 3s. by action       &              399 actions              \\
			        {2018}          &        EPIC-Kitchens        &    \cite{EPICKITCHENS2018}     &        70\%verb; 52\%noun; 41\%act.        & 40K seq. {11.5M im.}  total    &       125 verb, 331 noun       \\
			        {2017}          &        Kinetics-400         &  \cite{Kinetics2017_dataset}   &          61.0\%   (2-stream CNN)           & 400 seq. of 10s. by action     &              400 actions              \\
			        {2015}          &         ActivityNet         & \cite{ActivityNet_dataset2017} &               64.8\%   (CNN)               & 200 seq. for 648 h.            &              203 actions              \\
			         2014           &          Sports-1M          &  \cite{Sports1M_dataset2014}   &               63.4\%   (SVM)               & 1M seq. total                  &              487 actions              \\
			         2014           &         {CGC-2014 }         & \cite{SuperVector-ECCVW-2014}  &              $*$ 79.1\% (SVM)              & 14K seq. total                 &               20 actions              \\
			         2013           &         {CGC-2013}          &      \cite{CGC2014trac3}       &              $*$ 72.9\% (SVM)              & 13K seq. total                 &               20 actions               \\
			        2011-12         &         {CGC-2012 }         &      \cite{CGC2012_eval}       &              $*$ 72.9\% (SVM)              & ({One-Shot}) Mono exp.         &               20 actions               \\
			         2012           &          {UCF-101}          &   \cite{ ConvBoVW_Wang2015}    &             91.5\%   (CNN+SVM)             & 3207 seq. $\sim$13K im.        &               24 actions               \\
			         2011           &          {CAD-60}           &     \cite{CAD60_Sung2012}      &               74.7\%   (HMM)               & 68 seq. $\sim${60K im.}  total &               14 actions               \\
			         2011           &          {HMDB-51}          &   \cite{ ConvBoVW_Wang2015}    &             65.9\%   (CNN+SVM)             & 101 at least per action        &               51 actions               \\
			         2010           &           UCF-50            &     \cite{UCF50_best_2013}     &                   92.8\%                   & 100 to 150 by action           &               50 actions               \\
			         2010           &           Olympic           & \cite{ ImprovedTraj_IJCV2015}  &                   90.4\%                   & 50 by action                   &               16 actions               \\
			         2009           &         Hollywood-2         &  \cite{ImprovedTraj_IJCV2015}  &                   66.8\%                   & 61 to 278 by action            &               12 actions               \\
			         2008           &       UCF-11-youtube        &      \cite{HMDB_dataset}       &                   84.2\%                   & 100 per action                 &               11 actions               \\
			         2008           &         UCF-sports          &      \cite{SurveyRGB2014}      &                   93.5\%                   & 14 to 35 by action             &               9 actions                \\
			         2008           &          Hollywood          &       \cite{Laptev2008}        &              56.8\%    (SVM)               & 30 to 140 by action            &               8 actions                \\
			         2005           &          Weizmann           &      \cite{SurveyRGB2014}      &                   100\%                    & $\sim$100 actions              &               10 actions               \\
			         2004           &             KTH             &      \cite{SurveyRGB2014}      &              97.6\%    (KNN)               & 789 actions total              &               6 actions                \\ \hline\hline
			                                                                         \multicolumn{6}{c}{\small Multi-view video streams}                                                                          \\ \hline
			         2010           &           MuHaVi            &       \cite{MuHaVi2010}        &                   58.3\%                   & 8 cameras and 14 actors        &               17 actions               \\
			         2009           &           i3DPost           &      \cite{SurveyRGB2013}      &                    80\%                    & 8 cameras and 8 actors         &               11 actions               \\
			         2009           &         HumanEva-2          &    \cite{HumanEva_dataset}     &                   84.3\%                   & 7 cameras and 4 actors         &               6 actions                \\
			         2006           &            IXMAS            &      \cite{SurveyRGB2013}      &                   80.4\%                   & 5 cameras and 13 actors        &               13 actions               \\ \hline\hline
			                                                                         \multicolumn{6}{c}{\small From fixed action images}                                                                          \\ \hline
			         2013           &      89-action-dataset      &    \cite{89Action_Dataset}     &                   74.2\%                   & 2038 im. total                 &               89 actions               \\
			         2012           &          VOC-2012           &     \cite{Everingham2015}      &              69.1\%    (SVM)               & 400 im. min. by action         &               10 actions               \\
			         2011           &          VOC-2011           &      \cite{Everingham10}       &              64.1\%    (SVM)               & 200 im. min. by action         &               10 actions               \\
			         2011           &     {Stanford-40-Act.}      &  \cite{StanfordDataset_2011}   &                   45.7\%                   & 180 to 300 im. by action       &               40 actions               \\
			         2010           &          VOC-2010           &      \cite{Everingham10}       &                   62.2\%                   & 50 to 100 im. by action        &               9 actions                \\
			         2010           &            PPMI             &    \cite{PPMI_dataset_2010}    &                    85\%                    & 300 im. by action              &               7 actions                \\
			         2010           &           Willow            & \cite{WillowHumaActions_2010}  &                  62.88\%                   & 911 fixed images               &               7 actions                \\ \hline\hline
		\end{tabular}
	}
	\label{tab:DatsetRGB}
	%\vspace{-0.15cm}
\end{table}
%\vspace{-0.35cm}

\subsubsection{RGB-based approaches for HAR}
 {One of the first successful approaches for HAR was the Spatio-Temporal Interest Points (STIP) proposed by Ivan Laptev in \cite{laptev2005space}  and its extension to the 3D temporal volumes (see Fig.\ref{fig:Harris3D}(a)) in \cite{Laptev2008}. In terms of feature vectors, these extractors produce {Histograms of Oriented Gradients} (HOG) and {Histograms of Optical Flows} (HOF) \cite{SC_HediTAbia2012}. Multiple works have relied on the HOG (see Fig.\ref{fig:Harris3D}(b)) and/or HOF descriptors to produce their low-level local features \cite{CAD60_Sung2012, CGC2011Article, CGC2014trac3, JMLR_Wan2013, MonnierCGC2014, GMM_DTW}. }
 
  	\begin{figure}[!t]
  		\centering
  		%\subfigure[\ ]{\includegraphics[width = 0.32\textwidth]{./figures/Harris2D.png}}\label{fig:harris2d}
  		\subfigure[\ ]{\includegraphics[height = 2.7cm]{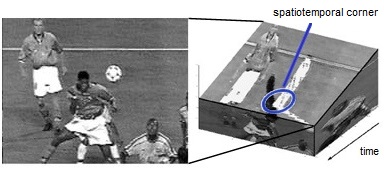}}\label{fig:stip} \ \ \ 	
  		\subfigure[\ ]{\includegraphics[height = 2.75cm]{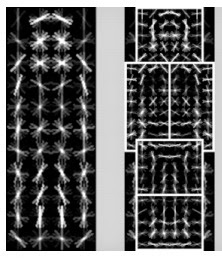}}\label{fig:hogDPM}
  		\subfigure[\ ]{\includegraphics[height = 2.7cm]{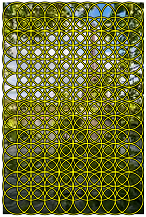}\label{fig:denseGrid}} \ \ \ 
  		\caption{\small { Example RGB-based descriptors: (a) Spatiotemporal interest points \cite{laptev2005space}, (b) HOG part-based  \cite{Felzenszwalb_2010} and (c) Dense grids \cite{DenseTraj_Wang2013}} }
  		\label{fig:Harris3D}
  	\end{figure}

 {The dense feature extractors shown in Fig.\ref{fig:Harris3D}(c) came later and added descriptors related to the motion boundary histograms and the motion trajectories \cite{DenseTraj_Wang2013}. The improved dense trajectories proposed in \cite{ImprovedDenseTraj2013} were based on a human detector to reduce trajectory features. They enabled one of the best literature performances in conjunction with the Bag of Visual Words (BoVW) sparse representation and the SVM-based classifiers \cite{SurveyRGB2014}. The combination of BoVW-based descriptors with CNN-generated features brought additional performance improvements \cite{FV_and_CNN2015}. This is partially explained by the capability of BoVW to encapsulate temporal informations complimenting the CNN spatial features. 
 	  	
 	A comparison between the BoVW Fisher vectors and CNN feature representations from large-scale data is given in Fig.\ref{fig:FV_vs_CNN}. According to the study of Guo et al. \cite{DeepSurvey2016}, the deep learning related approaches could be divided into 4 families: (i) Sparse-coding ones such as Fisher vectors, (ii) BoVW Auto-encoders and (iii) Restricted Boltzmann Machines (RBM) both of them (ii and iii) allowing to reconstruct the input data by automatically extracting (i.e. auto-encoding) the main features, in addition to (iv) CNN ones producing efficient feature representations using different architectures of convolution and pooling layers \cite{Molkanov_data_augment2016}.

	Actually, the main performance improvements are held by the approaches considering convolutional deep architectures to handle the actions temporal data-aspects. This factor is a major performance enabler compared to spatial-only approaches. This is confirmed by the study of Ghadiyaram et al. \cite{2020Trend} as learning masses and labels are constantly increasing and leaning towards weakly-supervised conditions. 
	The incorporation of the temporal evolution of continuous visual streams is affordable using a number of neural architectures: (i) Separately within the RNN or LSTM neural networks as in \cite{Pose_Zisserman2016, AGC_LSTM_Joint19}, (ii) using 3D convolutional kernels that exploit volumes of N consecutive frames and pixels \cite{Molkanov_data_augment2016}, (iii) by applying a cross-time pooling \cite{PigouCGC2016} or (iv) by combining the decisions of two-streamed networks, one of them relative to the optical flows \cite{ConvNets_2014}.
	An example approach that jointly extracts convolutional and spatio-temporal features is proposed in \cite{SpatioTemporal_CNN19}. By applying a weight-sharing constraint on the CNN-learned parameters, this method achieves the state-of-the-art performance on the Moments-in-Time challenging dataset \cite{MomentsTimedataset2018}.

 	\begin{figure}[!t]
 		\centering
 		\includegraphics[width=\textwidth]{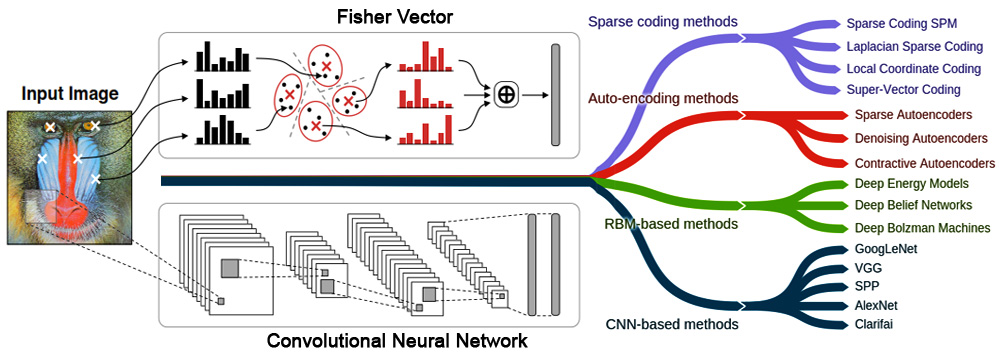}
 		\caption{\small {A comparison between the BoVW Fisher vectors and CNN large-scale representations \cite{FV_and_CNN2015} in addition to the relative deep methods families \cite{DeepSurvey2016} }}
 		\label{fig:FV_vs_CNN}
 	\end{figure} 
 	
	It is also worth mentioning that a number of recent approaches have focused on the use of different learning/testing conditions. Indeed, the number of frames or sequences per action can heavily influence the recognition performances \cite{2020Trend}. 
	This is for instance the case with the CAD-60 dataset \cite{CAD60_Sung2012}: Since each sequence can span on more than 4,000 frames, the mass of description data reaches more than 15K samples while the videos are limited to 68 sequences.
	In other learning scenarios, the use of {one-shot learning}, as in the CGC-2011/12 competitions \cite{CGC2011Article} creates different performance challenges as the test data is by-far bigger that the learning one. 
	Another category of approaches focused rather on ego-centric actions recognition (e.g. the EPIC-Kitchens challenge \cite{EPICKITCHENS2018}). In this case, having a camera attached to the person's head or chest, the recognition of hands motion plays a central role in actions recognition. These particular learning scenarios, as challenging they are, deserve to be further investigated in future works. In section \ref{sec:depth} dealing with depth-only approaches, we pay more attention to hands-dedicated estimation and recognition techniques.

\subsubsection{RGB-based datasets for facial expressions recognition}

 {The Facial Expression (FE) is a major information vehicle on human behaviors and emotions. This explains the race for developing many computer vision applications that include: Facial recognition/biometry, gender/age detection, 3D animation generation, FEs and emotions identification \cite{V.Bettadapura_2012}. Because a very-large population of datasets and contributions exist in relation to the mentioned applications, we focus more on those offering FEs as they are the most useful for SL visual recognition.}
 
 Within Tab.\ref{tab:DatsetRGBFaces} a classification of contributions is made chronologically for those having used unimodal images or video sequences. To the difference with body gestures, it is perceivable that many recent facial datasets (e.g. AffectNet \cite{AffectNet2017ValenceArousal}) are still exploiting fixed captures. On another level, when considering the labels column in Tab.\ref{tab:DatsetRGBFaces}, another classification could be made to 4 families: (i) Those using basic expressions \cite{FERET_best2017, ExpW_dataset}, (ii) those focusing on Action Units (AU) micro-expressions \cite{CKFace_2010, EmotionNet2016}, (iii) those considering facial pose estimation \cite{ExpW_dataset, MenPoDataset2017} and (iv) the more recent ones dealing with compound emotions under the Valence/Arousal evaluation system \cite{LIRIS_ACCEDE, AffWildDataset2019}. In addition to these FE recognition families, the active communities will be also described in what follows.

 \indent \textbf{Image-based expressions analysis:}
 The first image-based datasets offered increasing numbers of expressions. 
 They started with 2 expressions within the first benchmarking FERET dataset \cite{FERET_best2017}. It was the first multi-session dataset with 14K captures and 1199 subjects. 
 Within the next $AT\&T$ dataset, known as ORL, 4 expressions could be found in addition to occlusion with glasses \cite{FacialExp_survery2015}.
 Later, 5 different expressions were offered in YALE dataset \cite{YaleFaceDB}, to finally reach all 7 basic expressions (see next section \ref{sec:FE}) within the Japanese female faces dataset, i.e. JAFFE \cite{JAFFE_1999}. 
 
 Many next coming image datasets offered richer learning sets.  
 The Multi-PIE extended its previous CMU-PIE version to become the first multi-view (15 cameras) benchmark with 5 expressions and additional 19 illuminations \cite{Multi_Pie2008}.
 % relative to 337 subjects for a total of 750K acquisitions .
 The Static Facial Expressions in the Wild (SFEW) dataset, extracted from its dynamic version AFEW, was the first to offer challenging in-the-wild expressions \cite{AFEWdataset2011}. The training population continued to grow in the FER-2013 \cite{FERCdataset2013} and ExpW \cite{ExpW_dataset} datasets, to reach 36K then 92K faces, respectively. 
 
 \indent \textbf{Video-based emotions:}
 Starting from 2005, the first video datasets made their appearance with MMI-DB \cite{MMI_DB2005} and the extended Cohn-Canade (CK+) dataset offering 523 sequences and 23 AUs \cite{CKFace_2010}. 
 At this level, the dynamic analysis of AUs and emotions started receiving dedicated attention. 
 This was the case with AFEW \cite{AFEWdataset2011} with 1156 in-the-wild sequences, then with the AM-FED dataset with 6 AUs relative to spontaneous expressions from webcam recordings for a total of 168k frames \cite{AMFEDdataset2013}. 
 The extended CASME-II dataset offered on its turn 3000 videos, 247 micro-expressions and focused on high temporal resolutions (200fps) \cite{CASME2013_faceDB}. 
 DISFA+ is one of the latest improvements with stereo-videos particularly useful for analyzing the finest temporal AU changes within posed and spontaneous expressions \cite{DISFAdatasetFace2013}. 
 Nonetheless, the biggest AU dataset with 950K in-the-wild captures is held by EmotioNet \cite{EmotionNet2016}.
 
 \begin{figure}[!t]
 	\centering
 	\subfigure
 	{\includegraphics[width = \linewidth]{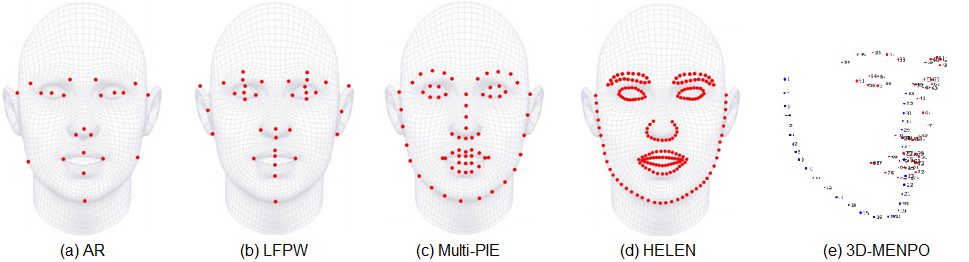}	} 	
 	\caption{\small Examples of facial landmarks: (a, b, c and d) 2D \cite{300VWdataset2016} and (e) 3D datasets \cite{MenPoDataset2017}}
 	\label{fig:FaceLandmarks}
 \end{figure}
 
\indent \textbf{Pose estimation:}
	From another point of view, other datasets allowed expressions analysis using pose estimation. Indeed, the facial pose vehicles information on both individual and group emotions \cite{ExpW_dataset}. For this purpose, increasingly richer benchmarks have offered different configurations of facial landmark points. As illustrated in Fig.\ref{fig:FaceLandmarks}, multiple 2D facial landmarks have been proposed \cite{300VWdataset2016}. The recent advancements are even considering 3D landmarks extraction from RGB-only faces \cite{MenPoDataset2017}. 

	Starting with 22 points within the AR dataset \cite{ARdataset1998}, the facial landmark annotations have increased to 29 points within the LFPW dataset \cite{LFPWFaceDataset2011}. Within the IMM image-dataset, 58 landmark points have been given \cite{FaceDb_IMM2004}. Then within 300-W and 300-VW dataset both 51 and 68 landmarks have been offered \cite{300VWdataset2016}. 
	And while the highest number of annotations is held by the Helen dataset with 194 points \cite{HelenDataset2012}, these positions could be interpolated from fewer numbers. 
	Within the recent 2D-MENPO video dataset, 68 points were considered from 12K frame samples. For the interesting 3D-MENPO dataset, the 84 landmark annotations were used for the first time in 3D from RGB-only faces \cite{MenPoDataset2017}.
 
 \indent \textbf{Compound emotions (Affect) analysis:}
	 If we continue in scale increase, we find that recent datasets were targeting the 1M frames in size and that they were ``forced" to consider compound labeling (e.g. disgusted-smile, sad-smile, etc.). 
	 The first attempt with 160 open movies and 73 hours has been offered by the LIRIS-ACCEDE dataset \cite{LIRIS_ACCEDE}. Next, came the RAF-DB dataset with an interesting set of 40 labels per FE. The selection of those with-consensus allowed the definition of 11 compound labels \cite{RAF_DB_face2017}.
	 On the top positions of Tab.\ref{tab:DatsetRGBFaces}, we find that the AffectNet dataset  \cite{AffectNet2017ValenceArousal} is actually the biggest image-based benchmark with 1M faces relative to 8 emotions. The biggest number of videos with 298 sequences and over 1.2M frames is actually held by the Aff-Wild dataset \cite{AffWildDataset2019}.  
	 While LIRIS-ACCEDE and RAF-DB used crow-sourcing for labeling \cite{LIRIS_ACCEDE, RAF_DB_face2017}, the AffectNet and Aff-Wild \cite{AffectNet2017ValenceArousal, AffWildDataset2019} where annotated by 2 and 6 experts, respectively. 

 \indent \textbf{Active communities:}
 Multiple recent competitions created by specialized communities exist and are open for new contributions.
 One interesting competition is the one organized by the MediaEval community. It has proposed multiple benchmarks using, among others, the LIRIS-ACCEDE large data collection \cite{LIRIS_ACCEDE}.
 The iBUG research group, on its side, contributed to the definition of multiple benchmarks in recent years \cite{ZafeirouSruvery2018}. Examples include: MMI-DB with 848 videos \cite{MMI_DB2005}, SEMAINE  with 959 human-agent discussion sequences \cite{SEAMINEdataset2012}, the 300-W and 300-VW pose estimation datasets \cite{300VWdataset2016}. 
 Active recent competitions include: (i) EmotiW  offering near 2K sequences for group level emotions recognition \cite{EmotiW2017}, (ii) the Aff-Wild  challenge  with 1.2M frames of random emotional states \cite{AffWildDataset2019}, and (iii) 3D-MENPO for facial pose estimation in real-world coordinates \cite{MenPoDataset2017}.

 %Interesting contribution paths are available through the different active competitions maintained by these specialized communities. 
  
  Within Tab.\ref{tab:DatsetRGBFaces}, we give a comparative ordering for the RGB-based datasets allowing FEs approaches. While many competitions have proposed iterative content improvements, we summarize by keeping the most representative recent versions (e.g. 300-VW instead of 300-W \cite{300VWdataset2016}) and indicate the references relative to the best respective performances.
  This list is continued under multi-modal acquisitions in section \ref{sec:MultimodalFaces} and Tab.\ref{FacialDatsets}.
 
 \begin{figure}[!t]
 	\centering
 	\subfigure[\ ] 
 	{\includegraphics[height = 2cm]{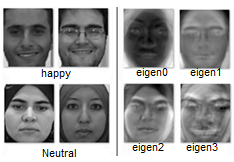}}	
 	\subfigure[\ ] 
 	{\includegraphics[height = 2.2cm]{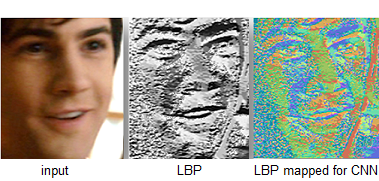}	} 	
 	\subfigure[\  ] 
 	{\includegraphics[height = 2.2cm]{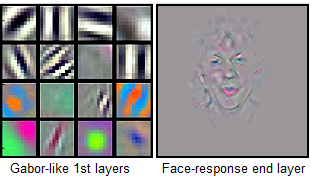}	}
 	\caption{\small Examples of commonly used facial features: (a) PCA-based Eigenfaces \cite{BSeddikSSD_2013},(b) Local Binary Patterns and their adaption to CNNs \cite{FaceExpSurvey2018} and (c) visualization of facial neural responses at different CNN layers \cite{GaborFilters2013, CNN_visulisation2019}  }
 	\label{fig:BasicFeatures}
 \end{figure}
 
 \begin{table}[!t]
 	\caption{\small Main RGB-based datasets allowing, among other uses, FEs recognition. (//) same configuration as above, ($\leftrightarrow$) The performance is relative to the tolerance  Margin-error considered}
 	\centering
 	\scalebox{0.8}{
 		\begin{tabular}{|C{1cm}|c|c|c|L{5cm}|c|}
 			\hline
 			\textbf{ \textsc{Year}} & {\textbf{\textsc{Dataset}}} &     {\textbf{\textsc{Ref. }}}      &  \textbf{\textsc{Classifier: Perf.}}  & {\textbf{\textsc{Content description}}} & {\textbf{\textsc{Labels}}} \\ \hline\hline
 			                                                                          \multicolumn{6}{c}{\small From  facial video streams}                                                                           \\ \hline
 			         2019           &          Aff-Wild           &     \cite{AffWildDataset2019}      &              CNN: 50.4\%              & 298 videos, 1.2M frames in the wild     &      Valence/Arousal       \\
 			         2018           &           RAVDESS           &     \cite{RAVDESS2018dataset}      &           75\% (video only)           & 26 subjects vocalizing with emotions    &         8 emotions         \\
 			         2018           &          Mobiface           &       \cite{lin2018mobiface}       &              CNN: 58.6\%              & 80 smart-phone streams, 95k frames      &       14 attributes        \\
 			         2017           &           EmotiW            &         \cite{EmotiW2017}          &        Many solutions: 80.9\%         & 1809 sequences of 5 sec. expressions    &       6 expressions        \\
 			         2017           &          3D-MENPO           &      \cite{MenPoDataset2017}       & CNN: 79.8\% ($\leftrightarrow$1.68\%) & 12K frames from many datasets           &       84 face 3D pts       \\
 			         2017           &          2D-MENPO           &      \cite{MenPoDataset2017}       & // 76.5\% ($\leftrightarrow$0.0024\%) & // (semi-frontal set)                   &       68 face 2D pts       \\
 			         2016           &           DISFA+            &    \cite{DISFAdatasetFace2013}     &          Multiple AUs rates           & 12 AUs for 27 subjects, 66 landmarks    &      42 face actions       \\
 			         2015           &  \scriptsize{LIRIS-ACCEDE}  &        \cite{LIRIS_ACCEDE}         &         Manual: 83.5\%/86.2\%         & 160 movies, 9800 $\sim$10s. seq., 73 h. &      Valence/Arousal       \\
 			         2015           &           300-VW            &      \cite{300VWdataset2016}       &  CNN: 75.2\% ($\leftrightarrow$0.06)  & 300 sequences, 218K faces                              &       68/51 face pts       \\
 			         2013           &          CASME-II           &      \cite{CASME2013_faceDB}       &              SVM: 63.4\%              & 3000 sequences, 247 micro-exp.          &       6 expressions        \\ 			         
 			         2013           &           AM-FED            &      \cite{AMFEDdataset2013}       &          Multiple AUs rates           & 242 webcams, 5268 subjects, 168k im.    &       22 pts, 6 AUs        \\
 			         2012           &           SEMAINE           &     \cite{SEAMINEdataset2012}      &     75\% ($\leftrightarrow$0.07)      & 150 subjects, 959 5mn. discussions      &      cognitive states      \\
 			         2011           &            AFEW             &       \cite{AFEWdataset2011}       &                59.2\%                 & 1156 seq., 113K im., gender             &        pose/6 exp.         \\
 			         2010           &             CK+             &         \cite{CKFace_2010}         &            SVM+AAM: 94.5\%            & 593 sequences, 23 AUs                   &       6 expressions        \\
 			         2005           &           MMI-DB            &         \cite{MMI_DB2005}          &              SVM: 92.3\%              & 848 AU videos and 740 images            &        79 emotions         \\ \hline\hline
 			                                                                            \multicolumn{6}{c}{\small From fixed face images}                                                                             \\ \hline
 			         2017           &          AffectNet          & \cite{AffectNet2017ValenceArousal} &              CNN: 55.5\%              & 1M faces, 8 emotions                    &      Valence/Arousal       \\
 			         2017           &           RAF-DB            &       \cite{RAF_DB_face2017}       &        CNN: 74.2\% and 44.6\%         & 30K faces x 40 compound labels          &      7/11 expressions      \\
 			         2016           &         EmotionNet          &       \cite{EmotionNet2016}        &              CNN: 75.5\%              & 950K images , 12 AU rates               &           12 AUs           \\
 			         2015           &            ExpW             &        \cite{ExpW_dataset}         &         CNN: 70.2\% (subset)          & 92K faces in the wild                   &       7 expressions        \\
 			         %2013          &            IMFDB            &        \cite{imfdb_dataset}        &                   -                   & 35K im., 100 subj., age, illum.         &      4 poses, 6 exp.       \\
 			         2013           &          FER-2013           &       \cite{FERCdataset2013}       &                71.2\%                 & 36K facial images                       &       7 expressions        \\
 			         2013           &            CASME            &      \cite{CASME_BestResults}      &                60.82\%                & 195 micro expressions                   &        35 emotions         \\
 			         2012           &            HELEN            &      \cite{HelenDataset2012}       &         Euclidean min. dist.          & 2330 high-resolution images             &          194 pts           \\
 			         2011           &            SFEW             &       \cite{AFEWdataset2011}       &              CNN: 48.6\%              & 700 facial images                       &       6 expressions        \\
 			         2011           &            LFPW             &     \cite{LFPWFaceDataset2011}     &          SVM+Bayesian:  97\%          & 1432 celebrity faces                    &           29 pts           \\
 			         2008           &          Multi-PIE          &       \cite{Multi_Pie_Best}        &              SVM: 92.3\%              & 337 Subjects, 15 views, 19 illum.       &       5 expressions        \\
 			         2004           &             IMM             &       \cite{FaceDb_IMM2004}        &           AAM:  $\sim$100\%           & 240 face  images, 40 persons            &       58 pts, 3 exp.       \\
 			         1999           &            JAFFE            &         \cite{JAFFE_1999}          &         Multiple: $\sim$100\%         & 213 Japanese female images              &       7 expressions        \\
 			         1998           &             AR              &        \cite{ARdataset1998}        &                 99\%                  & 126 subjects, 4000 images, 2 sessions   &           22 pts           \\
 			         1997           &            YALE             &         \cite{YaleFaceDB}          &                99.3\%                 & 165 images, 15 subjects                 &       5 expressions        \\
 			         1994           &             ORL             &    \cite{FacialExp_survery2015}    &                 83\%                  & 400 images, 40 subjects, occlusion      &       4 expressions        \\
 			         1993           &            FERET            &       \cite{FERET_best2017}        &                 97\%                  & 14K images, 1199 subjects, 2 sessions   &       2 expressions        \\ \hline
 		\end{tabular}
 	}
 	\label{tab:DatsetRGBFaces}
 \end{table}
 	
\subsubsection{RGB-based approaches for FEs recognition} \label{sec:FE}

Our RGB-based FE approaches analysis will be pursued at three levels: (i) A chronological cover of FEs recognition evolution, (ii) a focus on CNN-based feature extraction and (iii) a listing of actual and future trends in the field. 

\indent \textbf{Development in expressions recognition:}
	One of the fundamental contributions is the {Facial Action Coding System} (FACS) of Ekman et al. \cite{Book_ekman1997face} as it defines the standard coding of facial movements.
	Many approaches use the basic FACS system composed of 7 expressions: `\emph{the neutral state}', `\emph{surprise}', `\emph{anger}', `\emph{fear}', `\emph{disgust}', `\emph{sadness}' and `\emph{happiness}'.
	The more advanced approaches consider the finer facial motions described within FACS using 44 basic AUs (for reminder: Action Units) relative to the muscles involved in the expression, in addition to other AUs for face and eye movements. For example, AU1 and 2 are relative to inner and outer brows raising, AU26 is for the jaw drop and AU45 is for eye blink \cite{Book_ekman1997face} 

 	For expression recognition, the first approaches exploited appearance-based projection methods such as PCA (see Fig.\ref{fig:BasicFeatures}(a)) and LDA (i.e. Linear Discriminant Analysis). 
 	The faces' eigenvectors are computed, then all learning images are projected within a reduced space to calculate their weights. The obtained parameters are used as face descriptors. 	
 	Multiple variants of these representations were tested on the first benchmarks (e.g. FERET and Multi-PIE datasets \cite{Multi_Pie_Best}).
 	A representative solution using them in combination with neural networks  is found in \cite{EigenNeuralN_2010}.
	A next improvement has been brought by the {Active Shape Models} (ASM) and {Active Appearance Models} (AAM) developed by Cootes et al. \cite{T.F.Cootes_et_al_2001}. Using a statistical model with controllable parameters, they have combined: (i) Geometrical deformable shapes with (ii) localized appearance representations such as PCA. 
	By modulating predefined deformable models, the ASMs and AAMs were first used for the assisted generation of 2D and 3D textured faces \cite{FaceMorph3D_1999}. Later, they were behind multiple solutions aiming at facial pose estimation  \cite{T.F.Cootes_et_al_2001, FaceDb_IMM2004}. 
	
	At this point, literature started considering the task of facial detection from non-cropped faces. Using 2D Haar features and a cascaded learning model, a successful facial detector was proposed by Viola and Jones \cite{ViolaJones_2001}. While detectors' robustness continued to improve using HOG and CNN features \cite{FaceExpSurvey2018}, they are still impacting the performance of recent approaches \cite{300VWdataset2016}. Challenging situations include small facial resolutions, occlusion, extreme poses and illumination variations. Additional preprocessing techniques such as facial colors normalization and pose alignment are actually applied to overcome these problems \cite{Survey_FacialExp_Multimodal2016}.
 	
    A hole set of approaches have also focused on the definition of more efficient facial descriptors. The Local Binary Patterns (LBP) have been extensively applied as they are illumination invariant and save the micro-texture variations of pixels neighborhood. Coupled with SVM classifiers, they have led to the best 63.4\% accuracy on CASME-II \cite{CASME2013_faceDB}.
    A recent combination of different LPB setups (i.e. orientations, thresholds) as an RGB image is shown in Fig.\ref{fig:BasicFeatures}(b). These maps served as an optimized input for multiple CNN learning paths leading to a 40\% performance gain on the EmotiW 2015 challenge \cite{LBP_CNN2015}. Other orientation-and-scale invariant descriptors such as HOG and SIFT have also been evaluated separately or in combination with LBPs on datasets such as SFEW, AFEW and RAF-DB \cite{AFEWdataset2011, RAF_DB_face2017}. 
    
\indent \textbf{From Gabor filters to CNN-based features:}	 
	Gabor filters figure also among the most used facial feature extractors \cite{GaborFilters2013, Survey_FacialExp_Multimodal2016}. When used in banks of different orientations and scales, they can encode both spatial and frequency dispositions of pixels intensities at different resolutions. Interestingly, it has been proved by Krizhevsky et al. in \cite{ImageNet2012} that the CNN first layers within the AlexNet architecture behave in Gabor-like filters when learned from a massive datasets such as ImageNet. 
	This has been confirmed by the visualization toolkit elaborated by Yosinski et al. using generative neural networks \cite{CNN_visulisation2019}. 
	As illustrated in Fig.\ref{fig:BasicFeatures}(c), the left-images synthesized to produce high activations for the first layer neurons, resemble greatly to Gabor filter banks. Accordingly, CNN layers gradually learn abstract combinations of these filters until reaching facial or emotion-specific specialized neurons, producing activations similar to the right-image synthesized in Fig.\ref{fig:BasicFeatures}(c) \cite{CNN_visulisation2019}. 
	
	In \cite{GaborNet2019}, a tentative for the replacement of CNN first layers by auto-generated Gabor filters has been introduced in the GaborNet model.
	The performance evaluation on the AffectNet FEs dataset has reported similar performances with a much faster model convergence. 
	More recently within the FACS3D-Net CNN model, a combination of 3D-CNN spatial features with LSTM temporal modeling has been successfully applied for AUs dynamic recognition \cite{FACS_3DNet2019}. 
	Another special CNN architecture known as the Hourglass model has been particularly successful lately both for face and body points estimation from RGB \cite{MenPoDataset2017}. Its symmetrical architecture resembles visually to a sand-hourglass: It is composed of a first down-sampling using pooling to a very-low resolution, then of an up-sampling by pixel-wise production. Different combinations of  this architecture have successfully applied stacking or cascading techniques to predict 2D poses then extrapolate 3D real-wold positions out of them \cite{hourglass_cascade2018}. 

\indent \textbf{Actual and coming orientations:}
	% NEW INPUTS AND USES
	In addition to movies and Internet in-the-wild frames, actual acquisitions are also considering smart-phone streams \cite{lin2018mobiface} and the finest facial motions such as mouth vocalization \cite{RAVDESS2018dataset, AffWildDataset2019}.
	%  RAVDESS  focuses on proposer mouth vocalization in combination with sound recording. The RGB-only method leads to 75\% accuracy.
	And beside individual's emotions, even interpersonal relations (e.g. friendliness, dominance, etc.) could be predicted using additional attributes such as gender, age and automatic pose tracking \cite{ExpW_dataset, MenPoDataset2017}. For what concerns facial pose tracking, as already initiated for the body \cite{densePose2018}, the next coming challenges will also consider dense points estimation in 3D world coordinates \cite{MenPoDataset2017}. 
	
	New proper annotation methods will be required for the latter tasks. This is also valid for the circular expression representation system controlled by the Valence and Arousal measures (a detailed analysis is provided in \cite{AffectNet2017ValenceArousal}). 
	As the use of compound-labels is shifting to continuous evaluation systems \cite{AffWildDataset2019, BodyEmotion2018}, labels locality preservation methods such the one used in \cite{RAF_DB_face2017} could bring partial solutions. 
	But the best will be to provide more annotators and labels consensus in future contributions. 
	A proof of this difficulty is found with the AffectNet dataset where the 2 expert annotators agreed only on 60.7\% of the labels due to emotions variations. 
	And while crowd-sourcing could provide more consensus, non-expert annotators might mis-understand the fine-details of AU labels as within the LIRIS-ACCEDE dataset \cite{LIRIS_ACCEDE}. 
	Thus, apart from data collection, proper procedures orienting the annotators to better labeling are still expected to improve.
 	
 	Finally, the application of previous FEs recognition approaches into concrete robotic platforms is among the most promising future exploitations. Initial tentatives for facial mimicking and autism therapy could be found within the SEER \cite{SEER_FacialRobot} and ENIGMA \cite{ENIGMA_autismRobot} projects, respectively.	

\subsection{Recognition from depth streams} \label{sec:depth}

	The problem of occlusion represents a major challenge for correct HAR. It is present in real-word scenarios, and specially those involving inter-person/object interactions \cite{MutiModalSurvey2018}. It causes considerable noise within the joint modality and hardens the body-parts segmentation within RGB streams. Besides using costly multi-view acquisitions, the depth modality represents an optimal solution for occlusion handling. This is logically explained by the extra Z dimensionality offered by depth as it simulates the human stereo vision system. A related analysis is proposed in what follows.

\subsubsection{Depth-based datasets for HAR}
For acquisition, there is a number of databases offering, among others, the depth modality using the Infra-Red (IR) sensors of Kinect V1 and V2 \cite{HON4D2013, HOPC_pointCloud_2014}. 
Many depth-only methods exploited the MSR-Action3D database \cite{ActionLet2012} within its $1/3$, $2/3$ or cross-validation configurations. 
Other works exploited the large-scale capabilities of the CAD and CGC benchmarks \cite{CGC2012_eval, CAD60_Cippitelli2016} reaching 12K actions sequences.
Another easier way to reach large scale depth datasets is their synthetic 3D generation as employed within Microsoft's MSR dataset \cite{Shotton2013PAMI}.
Different IR sensors were also used such as the Asus Xtion in \cite{DepthPose_FeiFei2016} or Intel's RealSense depth cameras in \cite{BigHand2Mdataset2017}. 

On another level, after a number of datasets dedicated for actions description \cite{ STOP_depth_2012, HOPC_pointCloud_2014, CAD60_Cippitelli2016}, we find that the recent trend is rather for pose estimation applications  \cite{DepthPose_FeiFei2016, BigHand2Mdataset2017, FPHABdataset2018}. 
Indeed, the need for better virtual-reality experiences has accelerated the production of datasets focusing on hand gestures in frontal and egocentric views \cite{HanPoseSurvey2019}. 
Examples include the MSR-Action-pairs \cite{HON4D2013} with 12 action labels and 360 sequences related to gaming scenarios. 
More recent pose estimation datasets include: (i) The BigHand2.2M with 2.2M depth images of hands and 21 joint position labels, (ii) the F-PHAB offering similar joints with 1.2K sequences, 105K frames and additional 45 action labels, and (iii) HO-3D offering 16 joint positions and 3 object models for the challenging task of coupled hands and objects pose estimation. 
A combination of these datasets is offered within the one-million-hands challenge \cite{camgoz2017subunets}.
\begin{figure}[!t]
	\centering
	\subfigure[\ ] 
	{\includegraphics[height = 2.2cm]{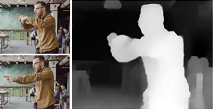}} \quad
	\subfigure[\  ] 
	{\includegraphics[height = 2.2cm]{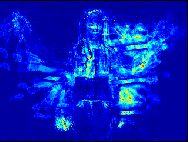}	}	\quad
	\subfigure[\  ] 
	{\includegraphics[height = 2.2cm]{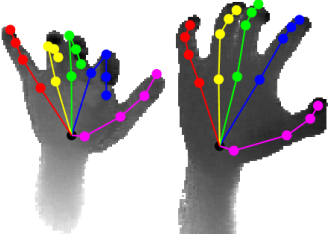}	}	
	\caption{\small Example depth-related approaches for: (a) Depth generation from videos \cite{DepthFromFrozen2019}, (b) motion-energy features extraction \cite{CGC2011Article} and (c) hand-joints pose estimation \cite{BigHand2Mdataset2017} }
	\label{fig:depthPose}
\end{figure}

\begin{table}[!tp]
	\caption{\small Main references and datasets allowing {depth}-based description/pose-estimation \cite{SurveyDepth2013, Survey3D2014, SurveyDepth3D2013}. ($*$) Adaptation as a percentage, (//) same condition as above}
	\centering
	\scalebox{0.8}{
		\begin{tabular}{|C{1cm}|R{4.2cm}|c|C{4cm}|c|c|}
			\hline
			\textbf{\textsc{Year}} &                                      \textbf{Reference} & \textbf{\textsc{Perf.}} & \textbf{\textsc{Content description }} & \textbf{\textsc{Labels}} & \textbf{\textsc{Dataset}} \\ \hline\hline
			\multicolumn{6}{c}{\small Spatial/temporal description from depth}                                                                            \\ \hline
			2018          &                           Garcia-H. {et al.} \cite{FPHABdataset2018} &         78.7\%          &      1.2K hand seq., 105K frames       &         45 acts.         &          F-PHAB           \\
			2016          &                      Cippitelli {et al.} \cite{CAD60_Cippitelli2016} &         93.9\%          &        68 seq. total, 4 actors         &            14            &          CAD-60           \\
			2014          &                           Tabia {et al.} \cite{HOPC_pointCloud_2014} &         84.9\%          &          $\sim$300 seq. total          &            30            &      UWA3D-multiview      \\
			2014          &                       Rahmani {et al.}  \cite{ HOPC_pointCloud_2014} &         96.2\%          &             336 seq. total             &            12            &      MSR-Gesture-3D       \\
			2013          &                                     Oreifej and Liu \cite{HON4D2013} &         98.2\%          &             360 seq. total             &            12            &     MSR-Action-pairs      \\
			2012          &                                  Guyon {et al.}  \cite{CGC2012_eval} &       $*$ 90.1\%        &              ({One-Shot})              &         8 to 12          &        {CGC-2012 }        \\
			2012          &                      Vieira {et al.}  ($1/3$) \cite{STOP_depth_2012} &         98.3\%          &             567 seq. total             &            20            &       MSR-Action3D        \\
			2012          &                     Vieira {et al.}   ($2/3$) \cite{STOP_depth_2012} &         96.8\%          &                   //                   &            //            &            //             \\
			2012          &                       Xiaodong {et al.}  ($CV$) \cite{DepthHOG_2012} &         91.6\%          &                   //                   &            //            &            //             \\ \hline\hline
			\multicolumn{6}{c}{\small Pose estimation from depth}                                                                                  \\ \hline
			2019          &       Li {et al.} ($\leftrightarrow$20px) \cite{DepthFromFrozen2019} &          41\%           &      2000 frozen actors sequences      &         21 pts.          &         Mannequin         \\
			2018          &     Hampali {et al.} ($\leftrightarrow$5mm) \cite{Ho3DdeptHands2019} &       $\sim$60\%        &     15K depth im., 8 pers, 3 obj.      &            16            &           HO-3D           \\
			2018          &   Garcia-H. {et al.} ($\leftrightarrow$14mm) \cite{FPHABdataset2018} &         69.2\%          &      1.2K hand seq., 105K frames       &            21            &          F-PHAB           \\
			2017          &   Yuan {et al.}   ($\leftrightarrow$5mm) \cite{BigHand2Mdataset2017} &          90\%           &     2.2M depth hand im., 10 pers.      &            21            &        BigHand2.2M        \\
			2017          &  Yuan {et al.}   ($\leftrightarrow$20mm) \cite{BigHand2Mdataset2017} &       $\sim$61\%        &      290K egocentric depth images      &            21            &            //             \\
			2016          & Haque {et al.} ($\leftrightarrow$10cm)   \cite{DepthPose_FeiFei2016} &         74.1\%          &      100K front/top depth images       &            14            &           ITOP            \\
			2012          &                      Ganapathi {et al.}   \cite{ganapathi12realtime} &          84\%           &        9K depth images, 24 seq.        &            8             &           EVAL            \\
			2011          &                            Shotton {et al.}   \cite{Shotton2013PAMI} &          77\%           &     900K synthetic/real depth im.      &            16            &            MSR            \\ \hline
		\end{tabular}
	}
	\label{tab:Depthdatasets}
\end{table} 

Another distinguishable category of datasets are those rather focused on varying the acquisition view-angles. One example includes the UWA3D-multiview \cite{HOPC_pointCloud_2014} with 300 sequences and 30 action labels. Another more recent example is the ITOP dataset \cite{DepthPose_FeiFei2016} applied for pose estimation with 100K front/top captures and 14 joint labels. 

Finally, we also highlight a totally different set of databases such as the Mannequin dataset \cite{DepthFromFrozen2019}. With 2000 RGB sequences containing fixed human actors, the goal is to reconstruct the depth maps from the intrinsic motion. While not directly related to action recognition or pose estimation, these datasets applied to depth creation from stereo-vision are full of promise  when applied to moving actors (see Fig.\ref{fig:depthPose}(a)) \cite{DepthMotion2018}. Within Tab.\ref{tab:Depthdatasets}, we highlight a selection of depth-only methods for actions recognition, pose estimation and their relative datasets.

\subsubsection{Depth-based approaches for HAR}

At the depth modality's level, although the apparition of the Kinect sensor in late 2010 enabled more related work \cite{Datasets_RGBD2016}, the number of research contributions found is still smaller than those using RGB or joint modalities. One reason might be the limited accessibility to this modality and the need to compensate for the noise generated by the infra-red sensors. A number of methods have also changed the data format from depth images to sets of 3D point clouds or 3D meshes after triangulation \cite{Blackburn_3D_Depth_2010}. Overall, the depth-based approaches could be divided in two families: (i) Those using the raw depth stream for feature extraction and (ii) those exploiting the displacement information for pose estimation.  
Both cases can enable HAR afterwards.

\indent \textbf{Depth-based actions description: }
{In terms of efficiency, we highlight that using the depth streams, a number of methods surpassed their equivalent in RGB \cite{survey2019small}. This is the case for example for the approach proposed in \cite{CGC2012_eval} within the CGC-2012 one-shot challenge. The best performance of $90.1\%$ obtained from the depth data was far above the $72.9\%$ accuracy obtained using the RGB streams.}
As features, a number of methods have focused on calculating energy-maps of displacements (see Fig\ref{fig:depthPose}(b)) at different frame instants from one action sequence \cite{STOP_depth_2012, CGC2012_eval}. Other features were also proposed using the 3D HOG descriptors \cite{DepthHOG_2012} or the histograms of discrete normals \cite{HON4D2013}. 
Similar features could also extracted from the reconstructed 3D point clouds: (i) A PCA-based representation was proposed in \cite{HOPC_pointCloud_2014}, (ii) an analysis of the 3D point clouds arrangements within well-defined networks has been also proposed in \cite{CAD60_Cippitelli2016}.
More recently, many approaches have migrated to CNN-based feature extractors: (i) A solution for the recognition of finger spelling using the basic CNN architecture AlexNet of Krizhevsky et al. \cite{ImageNet2012} has been proposed in \cite{fingerCNN_SL2015}, (ii) a more recent application of the 3D-CNN architecture \cite{DeepSurvey2016} to depth maps has also been adopted by Duan et al., as an improvement to previous BoVW methods in \cite{IsoGD_3Dconv2018}.

\indent \textbf{Depth-based pose estimation:}
 One advantage offered by the depth modality is the relatively-easy extraction of 3D joint skeletal positions compared to RGB. It was used under Microsoft's Kinect pose estimator proposed by Shotton {et al.} \cite {Shotton2013PAMI}. The authors relied on depth displacement differences and random decision forests to implement their real-time solution. Knowing that their algorithm was optimized for speed rather than accuracy, their resulting joint positions were relatively noisy and required additional post-processing \cite{Shotton2013PAMI}.
 A related improvement was proposed by Ganapathi et al. for the incorporation of skeletal and temporal constraints. Using the 3D point clouds generated from depth, they have used a probabilistic model for the body and the iterative-closest-point algorithm to improve the joints tracking \cite{ganapathi12realtime}. 
 
 Within the contributions dedicated to hands pose estimation, a similar extension could be found: A kinematic model defining the fingers pose relationships has recently been tuned within the MANO model \cite{Ho3DdeptHands2019}  composed of 21 joints (see Fig\ref{fig:depthPose}(c)). A successful incorporation of the kinematic constraints within the learning process of a deep CNN neural model has been also proposed in the HOLI-CNN model \cite{2016PoseKinematic}. To do so, the authors apply a hybrid approach composed of: (i) A cascaded CNN description for the partial joint poses, and (ii) a particle swarm optimization that integrates the kinematic constraints within the regression model. 
 
 The MANO and HOLI-CNN models were successfully applied to hands pose estimation on the Ho-3D and BigHand2.2M datasets. Within a 5mm error range, they have approximated 60\% and 90\% accuracies, respectively \cite{Ho3DdeptHands2019, BigHand2Mdataset2017}. 
 Under the more challenging egocentric views, we find that the HOLI-CNN estimator has been combined with LSTM networks for the 69.2\% action recognition performance obtained on the F-PHAB dataset \cite{FPHABdataset2018}. But, when using the egocentric subsample of the more challenging BigHand2.2M dataset, only 61\% performance was obtained within the relatively wide tolerance of 20mm \cite{BigHand2Mdataset2017}. This proves that more improvements are still expected.

\subsection{Unimodal temporal segmentation approaches}

When using spontaneous action sequences, the actions flow consecutively with minor resting momentums. The lexicon of recognizable gestures becomes then unlimited and correlated. The Temporal Segmentation (TS) step allows the decomposition of these continuous streams into manageable segments of more distinguishable actions. The TS results then in multiple $(start, end)$ frame couples delimiting the action positions. An example of the TS segments obtained for two moving hands is illustrated in Fig.\ref{fig:TemporalSegmentation}. We present in the following a classification of TS related approaches into 4 method families:

\begin{figure}[!t]
	\centering
	\includegraphics[width=\linewidth]{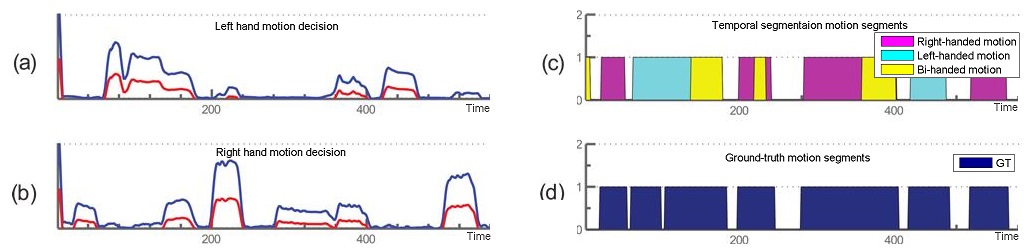}
	\caption{Example temporal segmentation results: (a) and (b) Left and right hands motion curves, (c) TS results and (d) the ground truth  \cite{ICIAP_SAGE2015}}
	\label{fig:TemporalSegmentation}
\end{figure}

\begin{enumerate}[i]
	\item \textbf{Heuristic:} 
	Using predefined thresholds, these methods allow Change-Detection (CD) \cite{CD_Roveri} from 1D signals (i.e. one computed feature). 
	They are then suitable for real-time decisions over time-series of data. 
	Peng \textit{et al.} \cite{SuperVector-ECCVW-2014} search for the most frequented body position with RGB frames, then extract the action segments having motions farther than a given $T$ threshold.
	Liang \textit{et al.} \cite{MotionTrail-ECCVW-2014} use rather the joint modality to implement a similar scheme and decide on the dominant hand in motion. 
	The CD approach presented by Seddik et al. \cite{ICIAP_SAGE2015}  uses the joint inputs to segment both the streams and the learning population into right, left and bi-handed actions (see Fig.\ref{fig:TemporalSegmentation}), thus allowing classifiers specialization in later steps. 

	\item \textbf{Dynamic programming:} 
	The commonly used techniques include the Dynamic Time Warping (DTW) \cite{DTW_humanActs} and Hidden Markov Models (HMM) \cite{UnifiedFramework2009}. In both case, the temporal slices are obtained by search for low-cost paths within generated weight matrices \cite{CGC2011Article}. The HMMs allowed top ranked performances in the Chalearn competitions requiring TS \cite{CGC2013DatasetResults}. Though DTW is a quite old technique \cite{DTW_spinger}, many approaches are still successfully using it \cite{JMLR_Wan2013}. A slightly improved DTW version using Gaussian Mixture Models (GMM) has been proposed in \cite{GMM_DTW}. A more recent DTW combination with deep CNN architectures is proposed by Junfu et al. in \cite{DTW_CVPR2019} for the purpose of continuous SL recognition.
		
	\item \textbf{Sliding windows:} 
	This method is used on a wider scale and consists of defining a number of frames (i.e. a window of $N$ seconds) used to segment and analyze the stream's features. This window is then slided with a predefined translation factor to browse the full stream \cite{SW_dynamic, CGC2013DatasetResults}. 
	Many approaches extract motion energy features \cite{MoHist} that encapsulate the temporal change differences. They can be extracted at different resolutions, from discrete grids of RGB, depth or directly from skeletal-joints positions \cite{SurveyLocGlob2010}. 
	
	\item \textbf{Learning-based:} 
	More advanced approaches are considering kernel-based or neural-networks-based solutions for the binary classification of frames into motion/non-motion \cite{KernelTS}. These classifiers could be optimized for accelerated executions \cite{TSrapidKernels2013} or for more robust segmentations as proposed by Neverova et al. in \cite{NeverovaPAMI2015}. More advanced approaches are combining RNN and CNN techniques to apply both the temporal segmentation and recognition within the neural architecture as proposed by Pigou et al. in \cite{PigouCGC2016}. This last configuration allowed the state-of-the-art performance on the challenging CGC-2014 benchmark.
\end{enumerate}

\section{Multi-modal methods}

 {Several recent approaches and databases are now profiting from multi-modal data availability. Many works benefited from 2D and 3D data affordable combinations. We focus here additionally on the approaches oriented towards static 3D actions/faces as-well-as dynamic ones. We summarize in what follows the details of these two research-work families.}

\subsection{Multi-modal datasets for HAR}\label{sec:mulmodalDatasets}
 {While the initial contributions combined the inertial motion sensors streams with RGB images \cite{HumanEva_dataset, CMU_MMAC2008}, the following advancements involved adding the actors' silhouettes masks \cite{MuHaVi2010} before giving place to a dominance of combined RGB, depths, and joint streams mainly coming from the Kinect sensor. This is the case for instance for the CGC and CAD benchmarks-based approaches \cite{CGC2011Article, CGC2014trac3, CAD60_Sung2012, CAD60_KoppulaGS13}. }

 {The most recent advances are considering learning on from large-scale data masses. This is for instance behind CGC-2011's dataset reformatting into the latest versions of ConGD and IsoGD 2016 datasets \cite{ConGD_IsoGD_cvpr2016} offering continuous and isolated action streams, respectively. This is also the case with one of the most recent and largest multi-camera dataset using Kinect-V2: NTU-RGB+D \cite{Kinect2_largeScale2016}.	
In the case of multi-view datasets, different viewing angles are combined to achieve higher performances. This is the case of Berkeley-MHAD, where 100\% of the actions are recognized \cite{MHAD_Berkeley2013} in a multi-modal and multi-views configurations. }

 {We summarize in Tab.\ref{MultimodalWorks}, the multi-modal contents for the most recent datasets. We also distinguish the solutions dealing with SL recognition and those offering continuous streams and therefore requiring a temporal segmentation stage before action recognition.}

\subsection{Multi-modal fusion approaches}
The combination of different modalities allows gaining their complementary informations for better performances \cite{Survey_Fusion2015}. This implies the use of different fusion strategies. The most used ones are detailed hereafter:

\begin{enumerate}[i]
	\item \textbf{Feature-concatenation fusion:} 
	The concatenation of low-level descriptors can be applied to combine multiple features coming from a unique modality \cite{ImprovedDenseTraj2013}.
	It can also be applied with different modalities such as RGB and depth to reduce feature variability \cite{JMLR_Wan2013}.
	Additionally, the selective extraction of the RGB and depth features to concatenate has been also applied using the joints as sparse position indicators \cite{CGC2014trac3, CAD60_KoppulaGS13}. In this special case where user-mask or joint positions are used for feature selection, high performance improvements have been reported \cite{Jhmdb_Act_Recog_Undesrstand2013}. This transition from low-level to mid and high level descriptors is illustrated in Fig.\ref{fig:ConditionalConcat}.
	Finally, full concatenation of the RGB, depth and joint descriptors has been also proven interesting \cite{Seddik_IET_2017}.

\begin{figure}[!t]
	\centering
	\includegraphics[width=\linewidth]{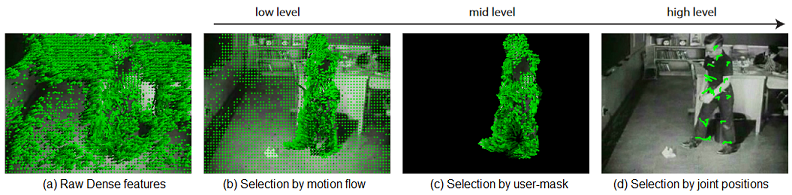}
	\caption{The selective concatenation of (a) raw Dense descriptors using (b) motion-flow, (c) user-mask and (d) joint positions to produce high-level features \cite{Jhmdb_Act_Recog_Undesrstand2013} }
	\label{fig:ConditionalConcat}
\end{figure}

\begin{table}[!tp]
	\centering
	\caption{\small Description of {multi-modal} datasets \cite{Multimodal_bianryStructure_2016}. `C': Color, `D': Depth, `J': Joints, `M': actor's Mask, `I': Inertial sensor. `Sign.': Relativeness to SL, `TS': Temporal Segmentation usage}
	\setlength{\tabcolsep}{3pt}
	\scalebox{0.8}{ 	
		\begin{tabular}{|c|c|c|cccll|l|c|cc|}
			\hline
			\textbf{\textsc{Year}} & \textbf{\textsc{Dataset}}       & \textbf{\textsc{Reference}}                     & \textbf{C} & \textbf{D} & \textbf{J} & \textbf{I} & \textbf{M} & \textbf{\textsc{Samples}}    &   \textbf{\textsc{Actions}}    & \textbf{\textsc{Sign.}} & \textbf{TS} \\ \hline\hline
			\multicolumn{10}{c}{\small Mono-angle view}                                                                      &  \\ \hline
			2019 & ASL-100    & \cite{ASL_RGBD2019}            & \checkmark & \checkmark &     \checkmark     & \_         & \_         & 15 subjects, 4K seq.      &       100  act.       &  \checkmark   &     \_      \\			
			2016 & ConGD and IsoGD    & \cite{ConGD_IsoGD_cvpr2016}            & \checkmark & \checkmark &     \_     & \_         & \_         & 21 subjects, 48K seq.      &       249  act.       &  \checkmark   &     \checkmark      \\
			2015 & UTD-MHAD          & \cite{UTD_MHAD_2015}                   & \checkmark & \checkmark & \checkmark & \checkmark & \_         & 8 subjects, 861 seq.       &       27  act.        &  \checkmark   &     \_      \\
			2014 & {CGC-2014}        & \cite{CGC2014trac3}                    & \checkmark & \checkmark & \checkmark & \_         & \checkmark & 40 subjects, 14K seq.      &       20  act.        &  \checkmark   & \checkmark  \\
			2013 & {CGC-2013}        & \cite{CGC2013DatasetResults}           & \checkmark & \checkmark & \checkmark & \_         & \checkmark & 20 subjects, 13K seq.      &       20  act.        &  \checkmark   & \checkmark  \\
			2013 & CAD-120           & \cite{CAD60_KoppulaGS13}               & \checkmark & \checkmark & \checkmark & \_         & \_         & 4 subjects, 120 seq.        &       10  act.        &      \_       & \checkmark  \\
			2013 & Berkeley-MHAD     & \cite{MHAD_Berkeley2013}               & \checkmark & \checkmark & \checkmark & \_         & \_         & 8 subjects, 861 seq.       &       27  act.        &      \_       &     \_      \\
			2013 & J-HMDB            & \cite{Jhmdb_Act_Recog_Undesrstand2013} & \checkmark &     \_     & \checkmark & \_         & \checkmark & $\infty$ subjects, 5K seq. &        51 act.        &  \checkmark   &     \_      \\
			2015 & BBC extended      & \cite{Pfister15}                       & \checkmark &     \_     & \checkmark & \_         & \_         & 40 subjects, 92 seq.       &     $\infty$ act.     &  \checkmark   & \checkmark  \\
			2012 & {CAD-60}          & \cite{CAD60_Sung2012}                  & \checkmark & \checkmark & \checkmark & \_         & \_         & 4 subjects, 68 seq.        &        14 act.        &      \_       &     \_      \\
			2011-12 & {CGC-2012}        & \cite{CGC2012_eval}                    & \checkmark & \checkmark &     \_     & \_         & \_         & {One-Shot}      &      8 to 12 act.      &      \_       & \checkmark  \\
			2012 & MSR-Daily         & \cite{ActionLet2012}                   & \checkmark & \checkmark & \checkmark & \_         & \_         & 10 subjects, 320 seq.      &        16 act.        &      \_       &     \_      \\
			2010 & MSR-Action3D      & \cite{MSRaction3D_2010}                &     \_     & \checkmark & \checkmark & \_         & \_         & 10 subjects, 557 seq.      &        8 act.         &      \_       &     \_      \\
			2010 & BBC               & \cite{Buehler10}                       & \checkmark &     \_     & \checkmark & \_         & \_         & 9 subjects, 20 seq.        &     $\infty$ act.     &  \checkmark   & \checkmark  \\ \hline\hline
			\multicolumn{10}{c}{\small Multi-cameras or multi-views}                                                                      &  \\ \hline
			2016 & NTU-RGB+D         & \cite{Kinect2_largeScale2016}          & \checkmark & \checkmark & \checkmark & \_         & \checkmark & 40 subjects, 57K seq.      & 3{$\times$}  60 act.  &  \checkmark   &     \_      \\
			2015 & UWA3D-multiview-2 & \cite{HOPC_pointCloud_2014}            & \checkmark & \checkmark & \checkmark & \_         & \_         &         \_               &  5{$\times$} 30 act.  &  \checkmark   &     \_      \\
			2015 & Berkeley-MHAD     & \cite{MHAD_Berkeley2013}				& \checkmark & \checkmark & \checkmark & \checkmark & \_		 & 12 subjects, 647 seq. 		& 4{$\times$} 11 act.	&	   \_ 		& 	  \_ 	  \\
			2014 & Human3.6M         & \cite{human36m_dataset2014}            & \checkmark & \checkmark & \checkmark & \_         & \_         &          \_              &        \_             &      \_       &     \_      \\
			2010 & MuHaVi            & \cite{MuHaVi2010}                      & \checkmark &     \_     &     \_     & \_         & \checkmark & 14 subjects, 136 seq.      & 8{$\times$}   17 act. &      \_       &     \_      \\
			2009 & Human-Eva-2       & \cite{HumanEva_dataset}                & \checkmark &     \_     &     \_     & \checkmark & \_         & 4 subjects, 56K im.        &        6 act.         &      \_       &     \_      \\
			2008 & CMU-MMAC          & \cite{CMU_MMAC2008}                    & \checkmark &     \_     &     \_     & \checkmark & \_         &           \_             &        5 act.         &      \_       &     \_      \\ \hline
		\end{tabular} 
	} %{\scriptsize \cite{ChAirGest2013}}
	\label{MultimodalWorks}
\end{table}

	\item \textbf{Score fusion:}
	%At the score level, the fusion is made out of modality-relative scores. 
	At this level, the fusion combines the scores produced by different unimodal classifiers. 
	They can be generated using probabilistic or discriminative decision models. In data complementarity is available, the fusion offers better performances than the separate models \cite{Survey_Fusion2015}. 
	Weighted fusion schemes have been applied from of multiple SVM classifiers' outputs in \cite{CAD60_Cippitelli2016, Seddik_IET_2017}. They have also been directly implemented within deep neural architectures in \cite{NeverovaPAMI2015, PigouCGC2016}.
	
		\begin{figure}[!t]
			%\vspace{-0.2cm}
			\centering
			\subfigure[]{	  	
				\includegraphics[width= 0.45\textwidth]	{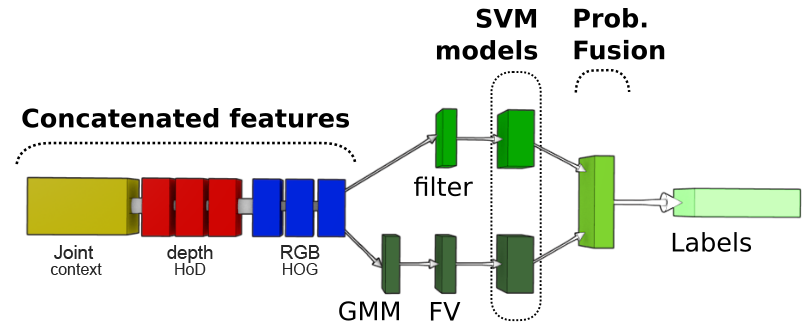} 
				\label{fig:fusionConcat}
			}	
			\subfigure[]{ 
				\includegraphics[width= 0.45\textwidth]	{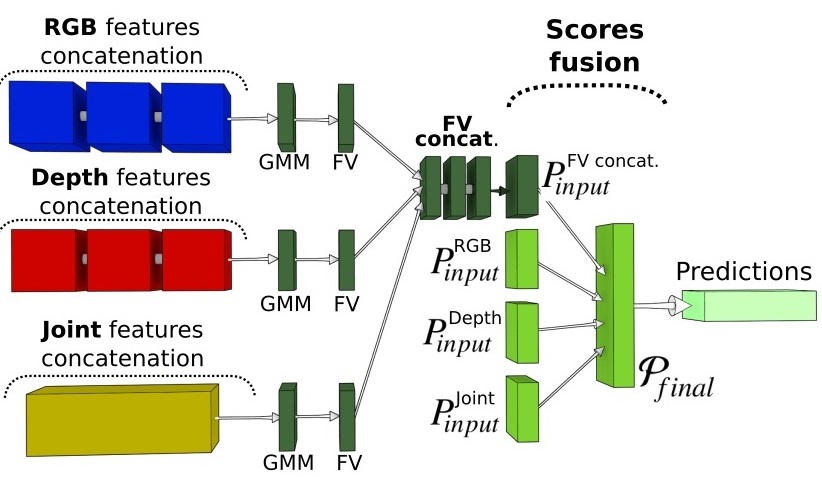}
				\label{fig:fusionResolution}
			}
			\caption{\small Examples of hybrid fusion schemes: Different concatenation and score fusions are applied in (a) and (b). Additional representation fusion is applied with FVs in (b) \cite{Seddik_IET_2017} }
			\label{fig:Fusions}
		\end{figure}
	
	\item \textbf{Representation fusion:}
	Peng et al. proved \cite{SuperVector-ECCVW-2014} that the fusion of different BoVW representations improves considerably HAR. The fusion process can be then improved using concatenation or pooling techniques to combine the different representations. 
	Within the CNN architectures, the fusion at the representation level is also considered a key part of the neural pooling process specially when using multi-modal inputs \cite{PigouCGC2016, DDNN_WuDi_CGC2016}.
	
	\item \textbf{Hybrid fusion:}
	The hybrid-fusion methods apply combinations of previous fusions as illustrated in Fig.\ref{fig:Fusions}.  
	Using a hybrid super-vector approach, advanced performances were obtained by Peng \textit{et al.} \cite{BoVWstudy_Peng2016} on multiple datasets by combining the decisions produced by multiple BoVW architectures.
	Similarly, in \cite{Seddik_IET_2017} the authors proposed a multi-layered fusion scheme of Kinect modalities for the recognition of SL actions from continuous streams. 
	Their solution starts by concatenating different modality-specific features, learns specialized SVM classifiers at local and global temporal scales, then optimizes their scores merging iteratively. %Figure \ref{fig:Fusions} illustrates different examples of hybrid fusion configurations. 

\end{enumerate}

\subsection{Multi-modal datasets for 3D FEs recognition}\label{sec:mulmodalFaces}

Before getting into multi-modal SL recognition, one special set of useful techniques are those exploiting 3D multi-modal facial data. 
%By holding 
These datasets can offer 3D meshes, point clouds, surface normals, IR maps in addition to RGB and landmark points.
%3D information , 
The acquisitions are then naturally less vulnerable to the frequent face alignment and lighting problems found with RGB-only methods \cite{FaceExpSurvey2018}.
They could be acquired using high resolution 3D scanners as for the Biwi-3D and the 3D-dynamic-DB datasets \cite{XMVTS2003, Boulbeba_base_visages_2014} or with multiple cameras and consumer depth sensors as within the Biwi-kinect \cite{Biwi_Kinect_3D} and the 4DFAB \cite{4DFAB_2018} datasets. 

A classification into static 3D or dynamic 4D facial datasets is offered in Tab.\ref{FacialDatsets}.
Many successful static datasets such as BU-3DFE \cite{BU3D_2006} and Bospherus \cite{Bosphorus2008} focused on fixed captures at peaks of expressions or AUs. 
Soon after, dynamic analysis of 3D FEs started gaining interest using, for instance, the extended versions of BU-4DFE \cite{BU4D_2008} and BP4D+ \cite{B4P2016} datasets.
An illustration of these benchmarks is given in Fig.\ref{fig:basesExpressions}.

In relation to pose estimation, many datasets also included 2D and/or 3D landmarks. This was the case with 6 points offered in 2D coordinates within Eurecom-Kinect dataset allowing a good separability by genre \cite{Eurecom_kinect_2014}. Other recent dynamic 4D datasets such as B4PD \cite{BP4Pdataset2014} incorporated 41 landmark annotations on top of both RGB images and 3D dynamic models. 

\begin{figure}[!t]
	\centering
	\subfigure[Bosphorus]{\includegraphics[height = 2.2cm]{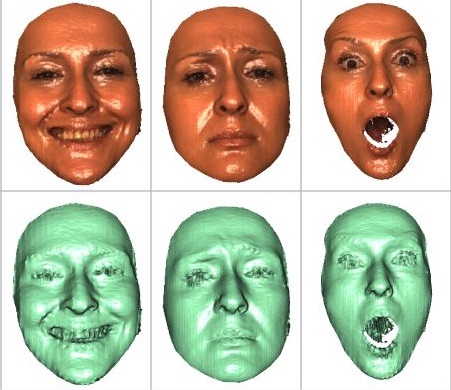}}
	%\quad
	\subfigure[BU-3DFE]{\includegraphics[height = 2.2cm]{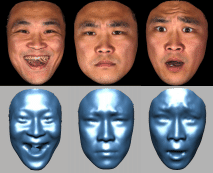}}
	%\quad
	\subfigure[BU-4DFE]{\includegraphics[height = 2.2cm]{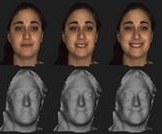}}
	%\quad
	\subfigure[BP4D]{\includegraphics[height = 2.2cm]{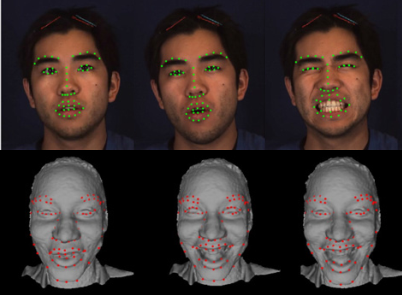}}
	\caption{\small Examples of 3D FE datasets: (a, b) Static and (c, d) dynamic}
	\label{fig:basesExpressions}    
\end{figure}

\begin{table}[!t]
	\caption{\small Static and dynamic 3D FEs datasets \cite{Survey_FacialExp_2012, Survey_FacialExp_Multimodal2016}}
	\centering
	\scalebox{0.8}{
		\begin{tabular}{|C{1.5cm}|C{2.5cm}|c|C{1.5cm}|L{7cm}|}
			\hline
			{\textbf{ \textsc{Year}}} & \textbf{ \textsc{Dataset}} &   {\textbf{\textsc{Reference}}}   & {\textbf{\textsc{Nb. Faces}}} & {\textbf{\textsc{Content description }}}            \\ \hline\hline
			                                                         \multicolumn{5}{c}{Approaches based on dynamic RGB+4D models }                                                          \\ \hline
			          2018            &           4DFAB            &         \cite{4DFAB_2018}         &              180              & 1.8M 3D meshes, posed and spontaneous videos        \\
			          2016            &           BP4D+            &          \cite{B4P2016}           &              140              & 10 emotions, 2D, 3D, thermal, physiological sensors \\
			          2016            &           BAUM-1           &     \cite{BAUM1emotions2016}      &              31               & 1502 emotions videos, frontal stereo + mono lateral \\
			          2014            &            BP4D            &      \cite{BP4Pdataset2014}       &              41               & Dynamic FACS emotions, 3D and 2D landmarks          \\
			          2014            &       3D-dynamic-DB        & \cite{Boulbeba_base_visages_2014} &              58               & Dynamic 3D faces, 15 fps, 4K vertex, pose free      \\
			          2012            &         Hi4D-ADSIP         &       \cite{Hi4D_faces2012}       &              80               & 3360 high-resolution sequences, 14 expressions      \\
			          2011            &          D3DFACS           &        \cite{D3DFAcs_2011}        &              10               & 100 3D frames per subject, 38 AUs                   \\
			          2008            &          BU-4DFE           &         \cite{BU4D_2008}          &              101              & 60K 3D frames, 606 sequences, 35K vertex par model  \\ \hline\hline
			                                                          \multicolumn{5}{c}{Approaches based on static RGB+3D models}                                                           \\ \hline
			          2017            &           MeIn3D           &         \cite{MeIn3D2017}         &             9663              & 12K 3D facial scans                                 \\
			          2016            &          FERG-DB           &  \cite{ExpressionCartoon3D2016}   &               6               & 55767 annotations on 3D cartoon emotion faces       \\
			          2014            &            KTFE            &   \cite{KTFE_termaldataset2014}   &              26               & Thermal+ RGB posed and spontaneous expressions      \\
			          2013            &       Eurecom-kinect       &    \cite{Eurecom_kinect_2014}     &              52               & Fixed RGB-D Kinect content with 6 annotated pts     \\
			          2013            &        Biwi-kinect         &       \cite{Biwi_Kinect_3D}       &              20               & Head pose and kinect RGB-D                          \\
			          2010            &            NVIE            &   \cite{NVIE_termalDatset2010}    &              100              & Thermal+ RGB 6 expressions                          \\
			          2010            &          Biwi-3D           &      \cite{ETH_Biwi3D_2010}       &              20               & Emotions and RGB-D                                  \\
			          2010            &          Texas-3D          &        \cite{Texas3D_2010}        &              105              & RGB-D; Facial recog. with 25 annotated pts          \\
			          2008            &          York-3D           &        \cite{YorK3D_2008}         &              350              & 3D content with textures                            \\
			          2008            &         Bosphorus          &       \cite{Bosphorus2008}        &              105              & 3D model with textures and 24 annotated pts         \\
			          2008            &            ETH             &          \cite{ETH_2008}          &              20               & Head pose and depth                                 \\
			          2007            &           CASIA            &         \cite{CASIA_2007}         &              123              & 4624 3D scans, $\sim$37 poses/illum./expression     \\
			          2007            &           Gavdb            &   \cite{Survey_FacialExp_2012}    &              61               & 427 3D faces, 3 expressions                         \\
			          2006            &          BU-3DFE           &         \cite{BU3D_2006}          &              100              & 6 3D facial exp., 100 subjects, 83 annotated pts    \\
			          2003            &           XM2VTS           &         \cite{XMVTS2003}          &              295              & speaking+rotating RGB sequences + a 3D model        \\ \hline
		\end{tabular}
	}
	\label{FacialDatsets}
\end{table}	

The largest set of static acquisitions is actually found within the MeIn3D dataset \cite{MeIn3D2017} with 9663 subjects and a total of 12K 3D scans. It has  served for the definition of the biggest morphable model till now. 
More recently, a 1.8M scale suitable for training deep-learning architectures has been provided for the first time under the 4DFAB multi-modal 3D dataset. Including 180 subjects, it holds spontaneous and posed expressions with applications related to both biometric and behavioral analysis.  
% THERMAL MUTIMODAL

Other interesting sets of 3D data include those considering IR thermal information as they are un-vulnerable to ambient lighting \cite{FaceExpSurvey2018}. Example datasets include the Natural Visible and Infrared facial Expression database (NVIE \cite{NVIE_termalDatset2010}), the Kotani Thermal Facial Emotion (KTFE \cite{KTFE_termaldataset2014}) and the richer 4D-dynamic BP4D+ dataset  with additional physiological sensors measuring blood pressure, heart and respiration rates \cite{B4P2016}.

\subsection{Multi-modal approaches for 3D FEs recognition} \label{sec:MultimodalFaces}

%R Vemulapalli, A Agarwala, “A Compact Embedding for FE Similarity”, CoRR, abs/1811.11283, 2018.
 {  We distinguish hereafter between static and dynamic multi-modal approaches.}

\indent \textbf{Static 3D approaches for FEs:}
	One of the first works to exploit the RGB+D for FEs recognition was proposed in \cite{Tsalakanidou_et_al_2003}. The features were extracted using PCA analysis from RGB and depth from the XM2VTS database, then nearest-neighbor classification was applied. The decision of belonging to a given class (e.g. happiness, fear, etc.) was obtained by a score fusion process multiplying 2 euclidean distances obtained from each modality \cite{Turk_Pentland_1991}. 
	
	After face cropping and alignment, a large number of contributions have also considered the landmark positions to compute distance measures of facial variations \cite{Survey_FacialExp_2012}.  These landmarks could be projected on RGB and depth maps from 2D coordinates, or directly exploited in 3D coordinates. 
	Frequency-based feature extractors such as Gabor filters  have  been successfully applied on top of these positions for localized expressions description \cite{Bosphorus2008}.
	Other baseline methods proposed extended versions of the LBP descriptors with the BU-3DFE and Eurecom Kinect datasets \cite{BU3D_2006, Eurecom_kinect_2014}. 
	An extension of RGB Multi-LBP maps \cite{LBP_CNN2015}, to a richer set of LPB maps deduced from 2D+3D combinations has been presented in \cite{2D3DfaceTensor2019}. 
	By applying tensor low-rank modeling, both dimensionality reduction and classification were made with 82.9\% and 75.9\% accuracies on BU-3DFE and Bospherus datasets, respectively.
	
	A different  set of methods relied rather on automatic discretization into facial curves or interest points \cite{Survey_FacialExp_2012}. This was the case with the SIFT-based features extractor proposed by Berretti {et al.} in \cite{Berretti_et_al_2011}. A wider comparison of different 3D facial datasets and approaches using BoVW techniques such as SIFT is found in \cite{FaceSurvey_WaelWerda2015}. A related contribution optimizing the space of possible gradients between 3D facial poses to generate a 3D facial morphing model is also presented in \cite{H.Li_et_al_2010}. 

	%One particular 3D dataset, the BU-3DFE static database has served for a wide set of approaches dedicated to 3D FEs recognition. 

	Logically, more recent approaches have been considering deep neural architectures. 
	Using accurate facial parts localization for the brows, eyes, nose and mouth, a fusion of 4 dedicated CNN paths has been applied in \cite{3DFE_partbasedRecog2018} leading to 88.5\% in expressions recognition accuracy on BU-3DFE.
	A fusion at the representation level of the CNN features produced from RGB and depth paths has been presented with an accuracy of 89.3\% on the same dataset in \cite{RGBD_fusionFace2017}. Due to noise in eyes regions, it has been shown that fusion of 2D and 3D facial CNN features improves performance. A fusion of CNN paths from depth, RGB, curvature and the 3 X/Y/Z normal maps has been proven efficient on subsets of the BU-3DFE and Bosphrus datasets in \cite{multiPathCNN_face2017}.
	More recently, data focused methods have also been successfully applied. Trimech et al. \cite{Hamrouni2017} reached 94.9\% accuracy by doubling the BU-3DFE dataset's size using data-augmentation with the coherent-point-drift algorithm and deep neural networks for learning and classification.

\indent \textbf{Dynamic 4D approaches for FEs:} 
	Classic methods rely on description of all data frames within a sequence, then on per-frame classification for a decision on the whole sequence label based on majority votes. BoVW-based methods offer additional robustness in this case as they are able to encode the expression using the most representative codebooks. Related to BoVWs, the FisherFaces representation allowed 82.3\% performance for the dynamic expressions recognition task on top of the Hi4D-ADSIP dataset \cite{Hi4D_faces2012}.	
	
	On top of the BP4D+ dynamic dataset, different other baseline methods have been evaluated separately on each modality. For instance, the use of Gabor features and ASM refinement led to 91.6\% 3D landmarks tracking (with 10\% tolerance error) on top of the thermal data of 60 random subjects. Using another similar random subset, 91\% was obtained using SIFT features and SVM classifiers for expressions recognition. 
	
	In addition to expressions description, many representation-based approaches (PCA in particular) are extensively used for  3D datasets analysis and the generation of 3D parametric models suitable for dynamic 3D animations. 
	By extracting the K prominent components representing the data, it is then possible to control the dynamic creation of synthetic 3D faces with their relative textures as in \cite{MeIn3D2017} or facial-animation blendshape transitions as in \cite{4DFAB_2018}.
	
	A representative solution  for dynamic FEs recognition and 3D avatars animation using Kinect streams has been proposed by Weise {et al.} in \cite{Weise_et_al_2011}. Their assisted learning algorithm prompts users to record 19 different FEs then the RGB and depth informations are combined within a Gaussian mixture model. Using expectation-maximization,  they were able to predict associations between the on-line streams and those recoded with robustness towards depth noise. 
	%A similar contribution based on semantic correspondence has been also proposed in \cite{KinectFaceReconstruction_2011}
	A similar solution for 3D avatar animation using the 7 basic FACS emotions has been proposed later in \cite{BSeddikSSD_2013}. It relied on the PCA representations with the advantage of automatically learning the expressions to avoid user input. 
	A more recent tentative to model 3D avatar's FEs using CNNs was proposed in \cite{ExpressionCartoon3D2016}. Though it used fixed 3D captures, it could easily be applied to dynamic 3D animations generation. %as in \cite{Weise_et_al_2011} and \cite{BSeddikSSD_2013}
	
	Recently, recurrent neural network of type LSTM have been encountering success for encoding the dynamic evolution of temporally spaced features. As features, the PCA representations of the deformations with regard to the neutral state have been used in \cite{4DFAB_2018}. They have led to a 70.1\% expressions recognition rate on the 4DFAB large-scale dataset.	

	Beyond this paper's scope, it is also interesting to consider voice combination with visual data for dynamic 3D FEs recognition. BAUM-1 is representative of a challenging set of recent benchmarks (e.g. RAVDESS in 2D \cite{RAVDESS2018dataset}) for understanding dynamic vocalization. Using multi-path CNN fusions, a combination of sound and 3D visual data, the best performance obtained is 54.6\% \cite{BAUM1emotions2016}. Thus, improvements are still possible.

\section{Main contributions related to SL recognition}

  { Sign Language research works can be divided in two categories: (i) The first family of works is dedicated to the exhaustive enumeration of all gestures related to a given SL vocabulary. In this context we find a wide range of vocabulary datasets offering limited instances by action \cite{SL_advances2008}. They are therefore rather oriented to the task of the data indexation. Their exploitation is then more often limited to web-based searches using keywords and meta-data \cite{ASLLVD2008}. (ii) The second category of works concerns those oriented for the task of action recognition. In this context, it is worth mentioning that technologies based on markers such as colored gloves or sensor-based hand wearables (e.g. cybernetic glove) \cite{ArSL_Gans2013} allow very good performances. Nevertheless, considering visual and marker-less gestures recognition, the task becomes more delicate. In this precise context related to computer vision, we find that the vocabularies become much smaller and that the number of samples per action are greedily increasing with the most recent datasets \cite{ConGD_IsoGD_cvpr2016}, thus allowing for deep-learning-based implementations.}

\subsection{SL datasets } 
 {In this context, we distinguish two types of datasets: Those designed mainly as an actions' dictionary for a given SL, and those rather suitable to visual actions recognition. We detail both types hereafter.}

\subsubsection{SL dictionary indexation datasets}

 Although a large number SL databases are already established, many are not suitable for computer vision recognition tasks. This is mainly due to uncontrolled acquisition conditions including inadequate labeling, reduced resolutions, variable view-angles, moving cameras, occlusion and noise \cite{SLRsruvey_2005}.
 %One of the first papers summarizing the literature dedicated to SL recognition is given by Ong and Ranganath in 
 Many dataset creation efforts were rather targeting the categorization and indexation of the different sign gestures. Consequently, their common exploitations were related to viewers-accessibly and SL education purposes. Some limited contributions tried to extract video descriptive features, but rather focused on searching videos by content similarities \cite{ASLLVD2008}. %and did not consider actions recognition.  
 %Their applications include assisted dataset content navigation and actions searching by similarities, signs learning and Interactive human-machine interfaces.
 %This explains why the majority of them have not received much interest from the computer vision community. 

 Some more advanced efforts have involved the generation of long signing phrases from predefined gesture lexicons \cite{SL_advances2008}. When exploiting computer avatars for this task, many interesting applications have been proposed: 
 \begin{enumerate}[i]
	\item  Under the ELAN project, a realistic 3D avatar performing American Sign Language (ASL) has been iteratively improved. Its last version offers realistic multi-person interactions and expressions \cite{ELAN_ASLavarat2016}. 
	\item Other recent solutions are exploiting stylized 3D avatars for SL synthesis and interaction. Successful examples include the Mimix and Turjuman applications related to ASL and Arabic SL (ArSL) \cite{MindRockets2019}.
	\item   The Hand-Talk solution is also offering a coverage for Brazilian SL (BrSL) \cite{handTalk2019}. As in \cite{MindRockets2019}, it recently extended its avatar to perform as a Web-based SL translator.  
	\item	Another contribution to the indexation of Tunisian Sign Language (TnSL) is presented by Bouzid and Jemni \cite{TNSL_BouzidKEJ16}. It combines 3D avatars with the written SL notation \cite{SignWriting, TNSL_2013}. 		
 \end{enumerate}
Interesting future extensions of the signing-capable avatars include their exploitation in social interactions and virtual-reality experiences as in  \cite{VRintractionMedical2016}.
 
\subsubsection{SL visual-recognition dedicated datasets } 

The analysis of Tab.\ref{tab:SLdatasets} shows that the databases dedicated to visual SL recognition have evolved from fixed images \cite{menasy_ASL} and MoCap recordings \cite{ChineseSL_gants2004} to include richer sets of isolated video sequences \cite{SL_advances2008}, long and continuous signing phrases \cite{CGC2014trac3} gradually towards deep-learning suitable scales \cite{PHOENIX2015, ChinesSL_CGC_2015}. Some datasets followed multiple enrichments for theirs contents and labels. This is the case for CGC \cite{ConGD_IsoGD_cvpr2016} and specially the RWTH-PHOENIX benchmark as one of the most cited for visual SL recognition. 
Actually, the richest SL-related content is held by the KETI dataset with near 15K video sequences acquired by stereo cameras and performed by 14 signers (12 of them are native). It offers labels for 524 signs and 127 landmarks points for body, hands and face. Figure \ref{fig:SLdatasets} illustrates example contents for the listed datasets. 

 From another point of view, we can distinguish different SL communities having been particularly active in recent years. 
 	A large number of advanced works have concerned the British Sign Language (BSL) \cite{Pfister15} from continuous or pre-segmented acquisitions relative to television streams \cite{Kadir04a}. 
 	 Similarly, many other works were dedicated to the ASL \cite{ASLLVD2008, PHOENIX2015}. The Dicta-Sign project is one of the representative research actions facilitating Web communication with native SL speakers \cite{DictaSign_2012}.
	 Promoted by the Chalearn gesture challenges,  many emerging works have exploited Italian SL (ISL) actions \cite{CGC2014trac3}.  
  	 %The CGC datasets offered a number of easily understandable signs that can also be found within other international SL lexicons \cite{SL_advances2008}. 
	 %The German SL with the SIGNUM and RWTH-PHOENIX multiple corpuses
	 %As for the Arabic Sign Language (ArSL) related dataset, we find that efforts for the creation of fixed image capture have been made in a reference project dating from 2001 \cite{menasy_ASL}. 
	 Different other datasets have concerned other SL lexicons \cite{SL_advances2008}. Example dictionaries include the ArSL \cite{ArSL_review2014}, German SL (GeSL), \cite{PHOENIX2015}, Chinese SL (CSL) \cite{ChinesSL_CGC_2015}, French SL (FSL) \cite{FSL_2015}, Greek SL (GSL) \cite{SL_ASL_2012}, Turkish SL (TSL) \cite{TSL_2013}, etc.
	 Our approaches analysis in next section will consider their relative contributions.
	 %But most of them were kept to limited vocabulary sets preventing their usage in real-world applications \cite{ArSl_Kinect2016}. }

	One interesting recent dataset for Tunisian SL has been published under the name Icharati-Sawti \cite{IcharatiSawti2019}. It targets 200 electoral concepts and has the specificity of offering long explanation phrases in Tunisian SL with text and voice transcripts. Interesting paths for contribution include: (i) visual signs recognition and (ii) the automatic learning of associations between text/audio annotations and visual streams, as proposed by Bowden et al. in \cite{BSL_TVannotate2009}. 	
	
	\begin{figure}[!t]
		\centering
		\subfigure[{\tiny RWTH-PHOENIX\cite{PHOENIX2015}}]
		{		{\includegraphics[height=0.175\linewidth]{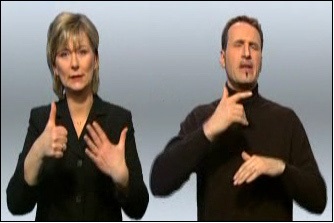}	}}
		\subfigure[{\tiny BBC-Extended\cite{PfisterPose2016}}]
		{		{\includegraphics[height=0.175\linewidth, width=0.22\linewidth]{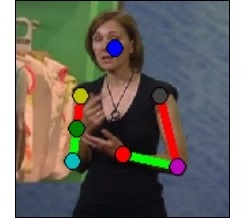}	}}	
		\subfigure[{\tiny Icharati-Sawti\cite{IcharatiSawti2019}}]
		{		{\includegraphics[height=0.175\linewidth]{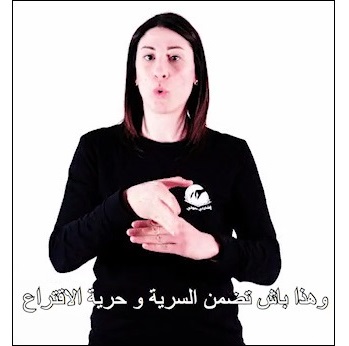}	}}
		\subfigure[{\tiny  KETI\cite{KETI_signLan2019}}]
		{		{\includegraphics[height=0.175\linewidth]{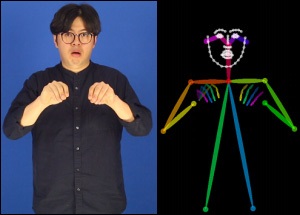}	}}					
		\caption{\small Examples of SL dataset contents in chronological  order. Notice the availability of different modalities in addition to RGB}
		\label{fig:SLdatasets}    
	\end{figure}
	
	\begin{table}[tp]
		\centering
		\caption[]{\small Datasets and communities dedicated to the visual recognition of SL \cite{SignLangFramework, SL_advances2008, SL_Spot2014}. $\infty$ : Continuous and uncontrolled signs. ($\star$) The 2 most trending datasets}
		\scalebox{0.8}{
			\begin{tabular}{|C{1cm}|L{5.3cm}|c|c|C{3.8cm}|}
				\hline
				\textbf{Year} & \textbf{\textsc{Research community: Dataset}}    &     \textbf{\textsc{Ref.} }      & \textbf{\textsc{Nbr. Sequences}} &            \textbf{\textsc{Nbr. Signs}}             \\ \hline\hline
				\textbf{2019} & \textbf{Korean sign language: KETI $\star$}              & \textbf{\cite{KETI_signLan2019}} &   \textbf{14,672 stereo seq. }   &   \textbf{524 by 14 signers + 124 face/body pts}    \\
				2019      & Tunisian sign language: Icharati-Sawti           &     \cite{IcharatiSawti2019}     &      200 electoral phrases       &                $\infty$ by 7 signers                \\
				2018      & South African sign language: Real-SASL           &       \cite{RealSASL_2018}       &    Multiple seq. per lexicon     &               800 on-line submissions               \\
				2016      & Arabic sign language: ArSL                       &      \cite{ArSl_Kinect2016}      &             400 seq.             &                         20                          \\
				2015      & British sign language: BBC-Extended              &         \cite{Pfister15}         &           92 long seq.           &                      $\infty$                       \\
				\textbf{2015} & \textbf{German sign language: RWTH-PHOENIX-2014 $\star$} &   \textbf{\cite{PHOENIX2015}}    &     \textbf{7000 seq. total}     & \textbf{1081 by 9 signers + $\sim$41 face/body pts} \\
				2015      & French sign language: FSL                        &         \cite{FSL_2015}          &            2500 seq.             &                         500                         \\
				2015      & Chinese sign language: Large-vocab.-SL           &     \cite{ChinesSL_CGC_2015}     &         3000 seq. total          &                        1000                         \\
				2015      & Chinese sign language: Daily-SL                  &     \cite{ChinesSL_CGC_2015}     &         1850 seq. total          &                         370                         \\
				2014      & Italian sign language: CGC-2014                  &       \cite{CGC2014trac3}        &          14K seq. total          &                         20                          \\
				2013      & Italian sign language: CGC-2013                  &   \cite{CGC2013DatasetResults}   &         13K seq.   total         &                         20                          \\
				2012      & Greec sign language: GSL                         &        \cite{SL_ASL_2012}        &            4920 seq.             &                         984                         \\
				2011      & Turkish sign language: TuSL                      &          \cite{TSL2011}          &             16 seq.              &                         29                          \\
				2011      & Multiples sign vocabularies: CGC-2012            &      \cite{CGC2011Article}       &           {(One-Shot)}           &                         249                         \\
				2010      & British sign language: BBC                       &         \cite{Buehler10}         &           20 long seq.           &                         100                         \\
				2008      & American sign language: ASLLVD                   &        \cite{ASLLVD2008}         &      $\sim$3000 seq. total       &                        1200                         \\
				2008      & British sign language: BSL-corpus                &      \cite{SL_advances2008}      &         8100 seq. total          &                         263                         \\
				2008      & German sign language: SIGNUM                     &      \cite{SL_advances2008}      &      19K seq., 780 phrases       &                  455 by 25 signers                  \\
				2004      & British sign language: BSL                       &         \cite{Kadir04a}          &         1640 seq. total          &                         164                         \\
				2004      & Chinese sign language: CSL                       &    \cite{ChineseSL_gants2004}    &            61K Mocaps            &                        5113                         \\
				2001      & Arabic sign language: Menasy                     &        \cite{menasy_ASL}         &        $\sim$1600 images         &                     $\sim$1600                      \\ \hline
			\end{tabular}  
		}  
		\label{tab:SLdatasets} 
	\end{table}
	
	We highlight in this context that the normalization of SL dictionaries through the creation of exhaustive visual datasets remains an open task for many languages \cite{PietroCelo2016}.
	This is specially the case as many SLs are actually mixtures of other lexicons (e.g. Tunisian SL contains among others ISL, FSL and ArSL \cite{TNSL_2013}). 
	Additional influencing factors include: (i) Vocabulary dissimilarities between younger and older aged signers, then (ii) geographic lexicon variations within a same country \cite{PietroCelo2016}.
	One solution that could help solving this problem is the call to crowd-sourcing for both data collection and labeling. This is the case with the RealSASL South African dataset \cite{RealSASL_2018}. On-line video submissions are accepted, then a validation procedure is applied afterwards by expert votes. They have been able to collect 800 sign video sequences from an initial dictionary composed of 6500 language lexicons \cite{RealSASL_2018}. 
	Applying this method is an interesting path for many other SL datasets.

\subsection{SL visual-recognition based works} 
 {Despite notable efforts by some research groups, the visual recognition of SLs remains less developed than the other branches of computer vision. Since the first steps engaged in this field under non-constrained (i.e. in the wild) and constrained experimental environments \cite{SL_advances2008, ASL_firstOnes1995}, this task encountered multiple challenges mainly related to body-parts segmentation and tracking. This is why the majority of the initial works focused on the use of colored and cybernetic gloves to circumvent members tracking problems. 

 Later, the apparition of multi-modal/depth-capable sensors (e.g. Kinect), in addition to the advances in RGB streams description \cite{CNN_visulisation2019} and in human pose estimation \cite{PoseEstimation_Shah2015}, allowed the development of interesting recent contributions. In what follows, we distinguish two categories of visual approaches: (i) The research-based contributions and (ii) the commercial solutions.
 	
\subsubsection{Research-works related to SL recognition}
 	As listed in Tab. \ref{tab:SLdatasets}, the SL visual recognition works can be grouped into specialized communities. We describe hereafter their main  contributions:}

 {\indent	\textbf{Works of Zisserman {et al.}:} Several contributions have been proposed by Zisserman and his team since 2004 for BSL recognition. A good number of them have focused on mono-modal recognition from television streams dedicated to the Deaf community \cite{Kadir04a, Buehler08, Pfister12}. 
 	%While this application relates to natural flows of native SL performers, 
 	%the basic challenges of 
 	These first contributions were kept unimodal and focused on body-members segmentation. 
 	Logically, the authors concentrated their next efforts in a number of recent works on the actor's pose estimation (generating the joints positions) using hands/arms images labeled with ground-truth positions. Thus, they benefited from RGB and joints modalities during actions recognition \cite{Pfister13, Pfister14a}.
 	Their latest advances followed the trend of large scale learning.  	
 	%This team has been interested recently in the pose estimation 
	This allowed their latest arms pose-estimator to reach an accuracy of 95.6\% within a 6-pixels reach from ground-truth \cite{PfisterPose2016}. They have used for this task a personalized CNN model that adapts to the occlusions found in each sequence.}

 {\indent \textbf{Works of Bowden {et al.}:} 
 	Working on their turn on multiple lexicons, Bowden {et al.} published many contributions related both to SL and HAR. 
 	 %before turning to the broader human actions in their recent productions. 
 	 In order to overcome the lack of large-scale databases dedicated to SL recognition, Cooper and Bowden proposed in \cite{BSL_limitedCorpora2009} an approach capable of learning within the special conditions of reduced dataset sizes or ground truth absence. 
 	 They have used pseudo-Haar volume filters to capture the sign actions from arbitrary temporal volumes. 
 	 An extension of this approach to partially labeled sequences has been proposed in \cite{BSL_TVannotate2009}. It exploited the textual annotations offered within SL television newscasts. 
  	 In their recent contributions, Bowden {et al.} incorporated CNN architectures for the recognition of localized actions relative to different body parts: The mouth in \cite{SLR_CNNmouth2015}, then the hands in \cite{Camgoz_2017_ICCV} and more recently for full SL translation \cite{SignLangBowden2018}. 
 	 
 	 For mouth shapes learning, they have forwarded the CNN soft-max outputs to an HMM model inspired from speech recognition models to reach 55.7\% as precision on a subset of 168K frames coming from the RWTH-PHOENIX dataset \cite{SLR_CNNmouth2015}. For hand-shape recognition, they have been able to propose the 80.3\% top-1 and 93.9\% top-5 state-of-art SubUNets architecture on the one-million-hands dataset \cite{camgoz2017subunets}.  Their approach relied on: (i) The CaffeNet CNN architecture for spatial description, (ii) on Bidirectional LSTMs (BLTSM) for temporal modeling and (iii) on a 12GB hardware GPU configuration that maximizes the frames loaded into GPU memory. One key contribution of BLSTMs is that they offer a better knowledge on the whole sequence by fusion of two opposite direction LSTM temporal layers. 
 	 
 	 Lower performances were obtained for continuous SL translation using SubUNets. For this purpose, they have proposed an improved approach that, as within Natural Language Processing (NLP) methods, accounts for grammar and words ordering for SL decoding. The evaluation process of their hybrid CNN-RNN-HMM model has reported 45.5\% and 19.3\% performance under the ROUGE and BLEU-4 respective scores on a large specifically-annotated subset of RWTH-PHOENIX \cite{SignLangBowden2018}. As there are many equally valid translations within a given language, the ROUGE and BLEU-1,2,3,4 measures are used within the newly emerging field of machine translation to describe the performance at different phrase scales \cite{SignLangBowden2018}.

 {\indent \textbf{{Chalearn} challenge related Works:} 
 	The Chalearn community has been very active in recent years for the creation of competition benchmarks related to SL actions, facial emotions and human behaviors \cite{cgcWebsite}. 
 	In its first CGC-2011 benchmark, the related competition proposed a large number of gestures belonging to multiple vocabularies 
 	(e.g. Taekwondo referees' actions, basketball, Indian SL, numbers, signs of divers, baby gestures,  etc.)
 	%Latin alphabet signs, acts of military soldiers, dance acts,
 	\cite{CGC2011Article}. The dataset offered limited numbers of gestures (around 10 for each vocabulary) made by non-native actors for one-shot learning goals. Thus, the research teams were forced to use 1 training sequence per action.} 
	
	In CGC-2013 and CGC-2014 datasets, the competitions proposed the first large scale human-actions database with 12K video sequences. The data inputs have been enriched to become multi-modal (from the Kinect sensor). The performance evaluation requested also the temporal segmentation from continuous streams \cite{CGC2013DatasetResults, CGC2014trac3}. 
	Later, the larger 2016 ConJD and IsoGD benchmarks combined formatted content from previous CGC datasets. In \cite{IsoGD_3Dconv2018} an IsoGD related contribution reached 67.3\% accuracy by combining the decisions of 3 different temporal and spatial CNNs for both the RGB and depth modalities. A better performance of 76.1\% has been reported in \cite{ASL_RGBD2019} using 5 CNN paths adding the hands and face extracted streams.
	Many other challenges, specially related to facial recognition applications are still actually active  within the Charlean community \cite{cgcWebsite}.
	%more on FEs and their contributions to the recognition of human actions. %This competition continuous till this moment of writing with competitions relative, among others, to Face Anti-Spoofing . cgcFace2019} 
% 	In terms of content richness for visual recognition purposes, the CGC competition dataset have been widely used in recent years \cite{cgcWebsite}. 

% {\indent	\textbf{ElMohandis {et al.} work:}
%A review of recent contributions to the recognition of Arabic Sign Language (ArSL) is given by ElMohandis {et al.} in \cite{ArSL_review2014}. A contribution to the visual recognition of the Arabic SL actions has been proposed in \cite{ArSL_FaceVisual2012} by means of a facial descriptors of Gaussian-based colors, geometrical metrics for description and Hidden Markov Models for classification. Many sensors experimentations have been also made using: Cybernetic gloves \cite{ArSL_Gans2013}, Leap Motion sensor \cite{ArSl_LeapMotion2014, ArSl_LeapMo2_2015} and Kinect \cite{ArSl_Kinect2016}. In most of these efforts, high performances were reported by means of SVM classifiers applied to various small databases. 
%These recognition rates are highly correlated with the acquired data that remains of relative small numbers and easy separability. 
%More interestingly, the ArSL collected datasets do offer a relatively exhaustive SL lexicon. }

 {\indent	\textbf{Other recent works:} 
 	The development of new solutions/devices assisting Deaf or hard of hearing people for SL communication is already receiving the attention from dedicated conferences such as {M-enabling} \cite{Menabling2019}. They allow SL promotion and the creation of services adapted to the concerned community.
 			
	At the spatial features level, having a set of joint positions for the body, face and/or hands, an open path for contribution is the extraction of high-level features from landmarks (i.e. joints) and their combination with the convolutional features given by CNNs. This is particularly the case for Liu et al. \cite{ConvD_J2016} merging the depth and skeleton body-relative modalities. A similar contribution is proposed by Jung et al. by combining the RGB and landmark facial modalities \cite{Face_CNN_landmark_2015}. Figure \ref{fig:TrendJointVisual} illustrates their respective contributions.}
	
	\begin{figure}[!t]
		\centering
		\subfigure[ ]
		{		\includegraphics[width=0.45\linewidth]{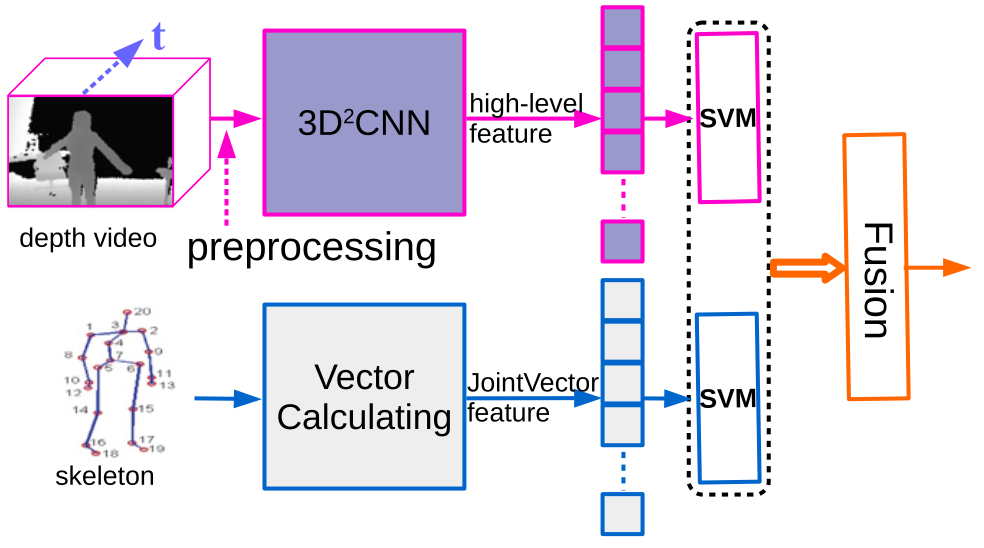}	}
		\quad
		\subfigure[]
		{		\includegraphics[width=0.45\linewidth]{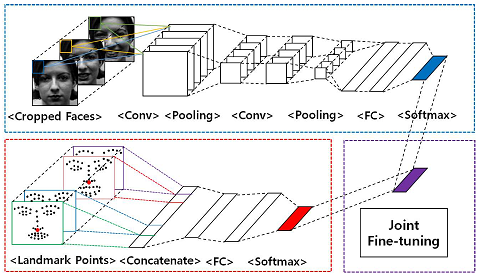}	}
		
		\caption{\small Example approaches merging skeletal (high-level) and spatial (low-level) modalities within deep architectures: (a) Body-related \cite{ConvD_J2016} and (b) face-related \cite{Face_CNN_landmark_2015} }
		\label{fig:TrendJointVisual}
	\end{figure}
	
	Many other recent research efforts have focused on modeling the temporal aspects of sign actions both for jointly their automatic description and segmentation. This could be made using a sequential pattern of interval descriptors and a decision-tree classifier as in \cite{SL_Spot2014}, or calling to newly specialized recurrent networks as in \cite{Molkanov_data_augment2016}. A solution avoiding temporal segmentation in the case of continuous streams is proposed in \cite{CSL_noTS2018}.
		
	The use of 2D/3D avatars for communication is additionally on of the very in-demand solutions for SL communication. Efthimiou {et al.} have proposed a recognition based on HOG/HOF descriptors in the first communication direction (Deaf $\to$ speaking person) and through the use of an avatar character in the reverse direction (speaking $\to$ Deaf person).  The recent advances as in \cite{stroySign2019} are offering increasingly realistic interactive characters.
		
	Finally, it is a fact that the combination of different modalities is a major performance improver. As recent methods are enabling the estimation of in-the-wild depth maps for the classically challenging non-rigid human bodies \cite{DepthFromFrozen2019}, there is place for major improvements in the actual human pose estimators \cite{OpenPose_2018, densePose2018}. As interesting are their relative results, the methods having exploited them are already indicating the need for next generation more efficient pose estimators \cite{KETI_signLan2019}. In this case, the progress made in the generation of depth maps can bring the required multi-modal boost in SL visual recognition performances.

\subsubsection{Commercial solutions related to SL recognition}

 %\indent	\textbf{Commercial solutions related to SL recognition :}  
 	The first successful SL recognition solutions focused on the use of sensors attached to the actor's upper body. The use of colored gloves or marker-based motion sensors on the face, body and fingers allowed effective and accurate motion tracking \cite{ArSL_Gans2013, stroySign2019}. They relied on multiple camera views with short and separate sequences of selected vocabularies. As such approaches require heavy hardware configurations, their applicability to real-world scenarios was kept limited \cite{SL_advances2008}. 
 	The main challenges encountered by commercial solutions are relative to the amount of sensors to exploit and the recognition effectiveness outcome. We present hereafter 4 active commercial solutions exploiting  visual informations for the creation of Deaf-dedicated services: 
	\begin{enumerate}[i]
		\item There is currently a solution called \emph{MotionSavvy} \cite{motionSavvy} targeting the American ASL users. It is based on the {Leap Motion} sensor packaged within a tablet interface (see Fig.\ref{fig:solutionsCommerciales}(a)). It allows both signs to text/voice translation (first direction) and voice to text recognition (second direction). Although taking advantage of the infra-red-generated depth stream of the sensor, it does not consider FEs. It also limits the actor's movements within a small space where both hands are accessible to the sensor. Therefore, it remains prone to future improvements.
		%while ambitious, are still experimental or in the early stages of development. 
		
\begin{figure}[!t]
	\centering
	\subfigure[\small{MotionSavvy}]{\label{fig:MotionSavvy}		
		\includegraphics[width = 0.235\textwidth]{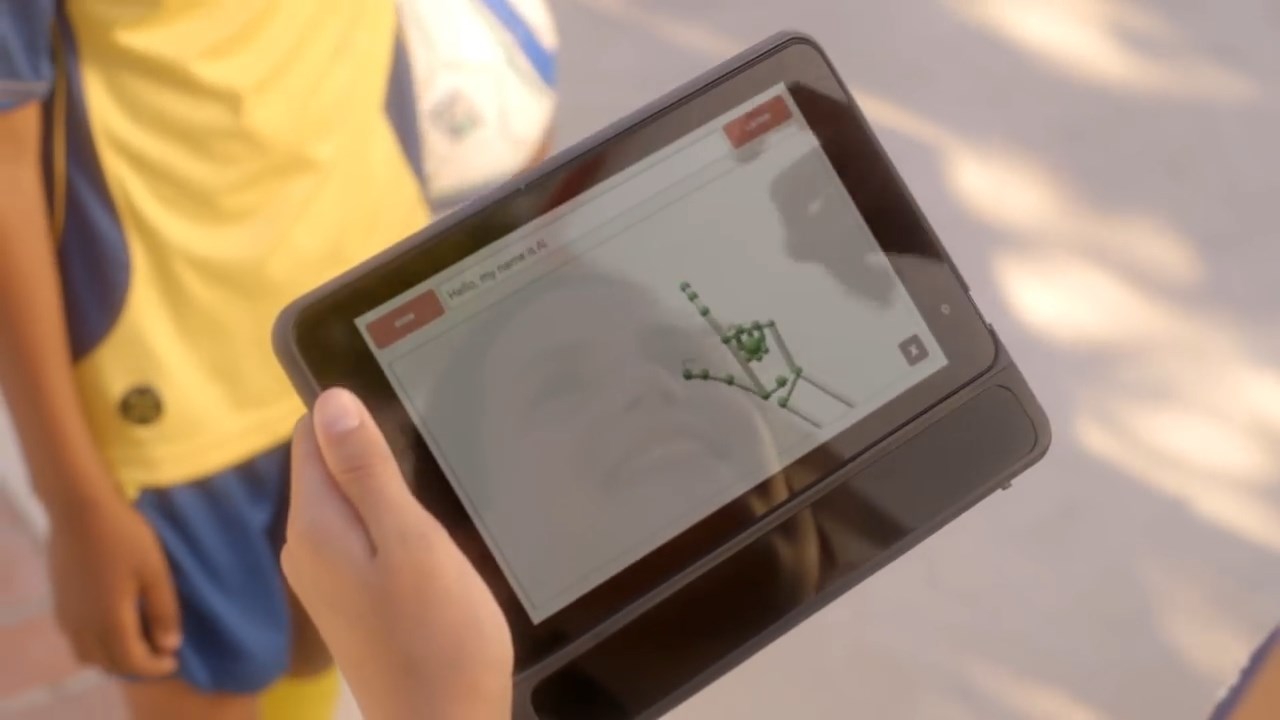}
		\includegraphics[width = 0.235\textwidth]{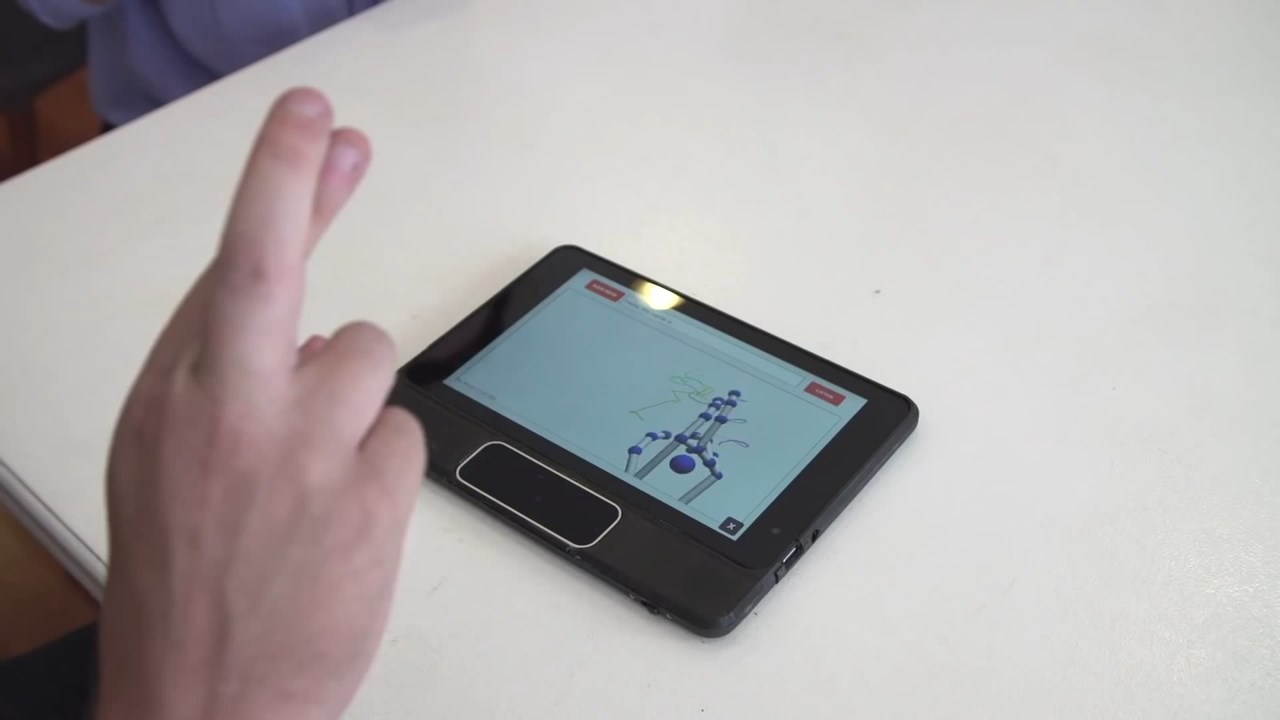}
	}
	\subfigure[\small{SignAll  }]{\label{fig:SignAll}		%PrimeSense et ASUS
		\includegraphics[width = 0.235\textwidth]{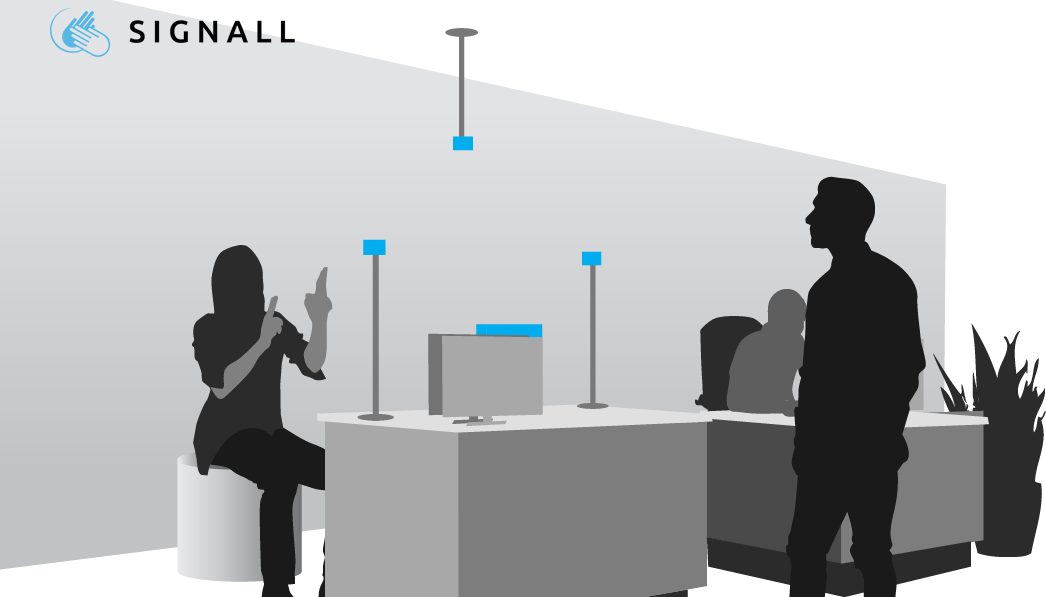}
		\includegraphics[width = 0.235\textwidth]{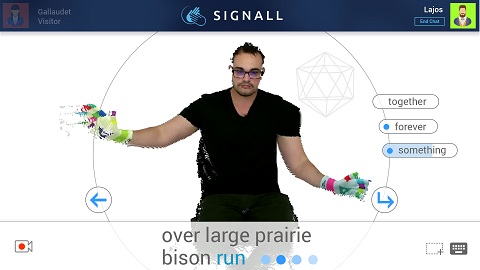}		
	}
	\\
	\subfigure[\small{OCTI  }]{\label{fig:OCTI}	%PrimeSense et ASUS
		\includegraphics[width = 0.47\textwidth]{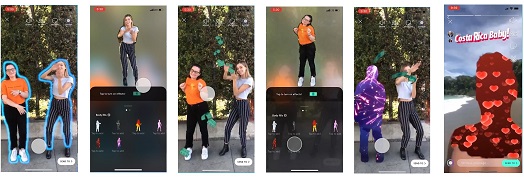}		
	} 	
	\subfigure[\small{StorySign  }]{\label{fig:StorySign}	%PrimeSense et ASUS
		\includegraphics[width = 0.235\textwidth]{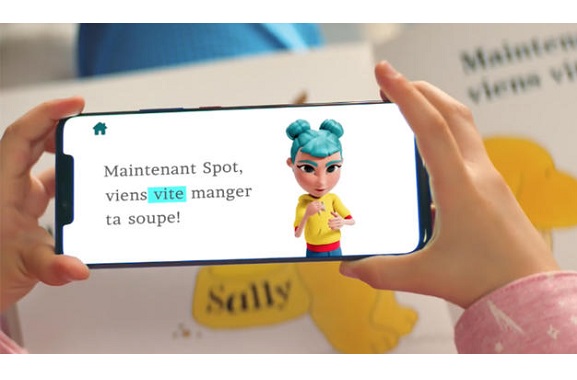}
		\includegraphics[width = 0.235\textwidth]{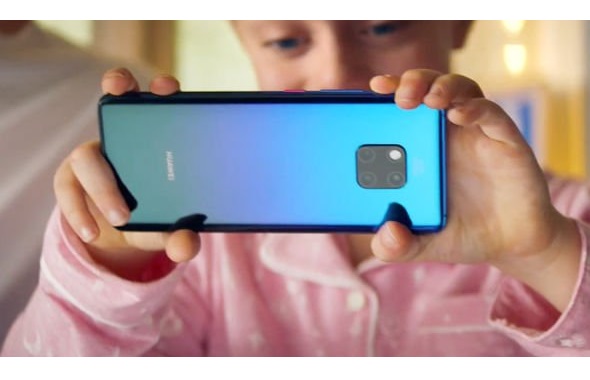}
		
	}	
	\caption{\small Examples of commercial solutions using gesture recognition}
	\label{fig:solutionsCommerciales}
\end{figure}
		
		\item {Another commercial solution based on the Kinect sensor is currently distributed under the \emph{SignAll} label \cite{SignAllApp2018}. It offers significantly more freedom for the actor, takes into account FEs and hand gestures through colored light-weight hand gloves. 
		Though its first iterations suffered from relatively limited performances, it holds an active development team. 
		Shown in Fig.\ref{fig:solutionsCommerciales}(b), its latest version exploits the improved Kinect-v2 in conjunction with multiple cameras, sound sensors and natural language processing for sentences generations. This product allows a Deaf agent to communicate through a terminal with speaking persons using ASL and Hungarian SL. Thus additional contributions are possible for other SL lexicons. }
			
		\item For what concerns RGB-only-based action recognition, the early solutions functioned within constrained environments. In this case, they required fixed and easily removable backgrounds. 
		%While, in real world scenarios, this task would be one of the first challenges to overcome. This is why the last generation of applications are focusing more on real-world scenarios. 
		Interestingly, new deep-neural based solutions such as the \emph{OCTI} application \cite{octi2018} (see Fig.\ref{fig:solutionsCommerciales}(c)) are starting to appear on last smart-phone generations to allow background removal and basic actions recognition for consumer visual affects. This new trend of smart-phone applications is expected to continue improving in coming years specially with the recent advances on body, face and hands poses estimation \cite{PoseEstimation_Shah2015, Pose_Zisserman2016, DepthPose_FeiFei2016, densePose2018}. %within the computer vision community.}
		
		\item % In this context, 
		One of the recent contributions joining the efforts of European, British and Australian Deaf organizations resulted in the \emph{StorySign} mobile application. It has been released in early 2019 in collaboration with Huawei AI \cite{stroySign2019} (see Fig.\ref{fig:solutionsCommerciales}(d)) to facilitate SL based education. Its originality consists in the use of state-of-the-art FEs recognition and motion capture technologies to allow high quality facial, hand and body performances by the 3D friendly avatar. As the amount of stories per SL are still limited within the \emph{StorySign} solution, this specific task of content creation and formating still remains wide open for new contributions.
	\end{enumerate}

\section{Conclusions and Discussions}

We have presented in the previous sections a review of HAR methods with consideration to their specific exploitation for visual SL recognition. We have followed a classification based on their input data-modalities. For each of the joint, RGB and depth input streams, we have listed the most used dataset contents and approach details. We have also accorded attention to the temporal segmentation task, in addition to the static and dynamic FEs recognition due to their importance in SL interpretation. Next, we detailed the new trend of multi-modal datasets and their relative contributions. 

In each section, we have been able to: (i) Discuss the solutions advantages, (ii) report the current best performances and (iii) highlight the most promising improvement paths. We finally payed special interest to the different working groups and commercial solutions targeting the Deaf and hearing-impaired community. 

We summarize hereafter the main contribution paths that we recommend at the levels of (i) datasets,  (ii) approaches and (iii) commercial solutions for the goal of SL recognition.
%the next promising approaches to consider 
 
\subsection{Datasets level}
From previous dataset listings, appears the need for multi-modal, large-scaled and SL-specific benchmarks. 
%It is clear that large scale dataset are lacking in the field of SL recognition. 
%It is not though specialized for a specific SL lexicon as it combines many versatile signs (e.g. sports, police, indian numbers, etc.).
To our best knowledge, the nearest multi-modal dataset to this goal is the KETI dataset proposed by Ko et al. in 2019 \cite{KETI_signLan2019} focusing on Korean SL and offering stereo RGB + landmarks streams. Though it relies on the OpenPose \cite{OpenPose_2018} estimator to generate its 127 landmarks, these positions allow only 55.3\% accuracy on the validation set for a limited selection of 105 sentences. Its creators already place as perspective the call to more robust pose estimators and the enrichment to larger-scaled recordings \cite{KETI_signLan2019}. The nearest multi-modal and large-scaled (13K sequences) dataset is CGC-2014. But it has the limitation of offering only 20 signs. Though many SL-dedicated RGB datasets might be suitable for large-scaled learning with appropriate data-augmentation techniques \cite{Molkanov_data_augment2016} (e.g. RWTH-PHOENIX dataset with 7000 sequences and 1081 sign labels), they remain either unimodal, or lack robust landmark annotations. Thus they remain open to contributions applying enriched formatting and labeling. This explains, for instance, the different extensions applied to the RWTH-PHOENIX dataset \cite{PHOENIX2015}. 

The design of any new SL dataset is preferred to be multi-modal, large-scaled and offering body, face and hands landmarks. Knowing that SL lexicons have thousands of gestures and follow different grammar-rules than spoken languages, the annotations should target the largest affordable lexicon coverage, both in isolated and continuous spontaneous streams. Also, as SL lexicons are constantly changing, it is preferable to record native SL signers after consensus on the dictionary to adopt. This was for instance the case for the relatively-small Tunisian electoral SL dataset proposed in \cite{IcharatiSawti2019}. Due to the logistic complexity of this task, specially if multi-view conditions were to considered as in \cite{Kinect2_largeScale2016}, this task remains a challenge to overcome by coming contributions. 

%\indent \textbf
\subsection{Approaches level}
After a relative prevalence for hand-crafted features in combination with BoVW representations, the capability of deep CNNs to learn automatic descriptive features from massive data offered a major performance leap in computer vision \cite{ImageNet2012}. From this perspective, a majority of community efforts have been oriented towards the collection of large-scaled datasets targeting actually the 1M samples and 1K labels \cite{AffWildDataset2019, AffectNet2017ValenceArousal, MomentsTimedataset2018, Sports1M_dataset2014}. This is an additional argument encouraging the creation of large SL-dedicated benchmarks as they will automatically lead to better performances using the actual CNN architectures. Next, we categorize the future promising improvements at the approach's level.

\indent \textbf{Streams preprocessing:}
Many intuitive contributions have successfully considered the preprocessing of input streams into more CNN-suitable formats to improve their accuracies. Example manipulations include the extension to a more-dimensional formatting (e.g. convolution along orthogonal views of volumetric videos \cite{SpatioTemporal_CNN19})   or the de-correlation of input modalities to a more easily-processable formatting (e.g. LBP descriptors for faces \cite{FaceExpSurvey2018}).

\indent \textbf{Multiple neural paths fusion:}
Considering that the incorporation of multi-modal fusion paths is a performance improver \cite{NeverovaPAMI2015, Seddik_IET_2017}, many recent contributions have combined two-streamed (and more) convolutional network paths for the description of both temporal and spatial features \cite{MutiModalSurvey2018}.
The two neural streams could fully or partially be dedicated to the description of action's evolution in time. This is the case with the dedication of one neural path for the motion flows in \cite{ConvNets_2014}.
The CNN paths could also be relative to different spatial modalities as employed by Duan et al. \cite{IsoGD_3Dconv2018} for the fusion of two 3D convolutional networks (i.e. from space-time volumes of pixels) one for RGB and the other for depth. 
Other joint-based contributions have also considered merging two LSTM paths: One for the joints description and the other for extracting  the kinematic relations in-between isolated joint groups \cite{AGC_LSTM_Joint19}. 
This can be seen as a deep neural equivalent to the Joint-Quadruplets \cite{jntQuadruplets} and the Moving-Pose \cite{MovingPose2013} hand-crafted features. 

\indent \textbf{Better neural description architectures:}
The last comparison reveals the actual orientation towards the creation of deep neural networks automating previously existing descriptors. Interestingly, even a spatiotemporal CNN-based equivalent to the widely used STIP descriptor of Laptev et al. \cite{Laptev2008} has been recently claimed in \cite{SpatioTemporal_CNN19}.
Some specific features such as motion trajectories, optical flows, joint shape contexts, etc. remained for a while absent within the CNN produced features. This resulted in a first set of contributions combining BoVW and CNN decisions for better performances \cite{ConvBoVW_Wang2015, ConvD_J2016}. 
Within next approaches, the temporal descriptors started to be modeled within the deep neural architectures using for instance RNN models and their optimized LSTM implementations. This allowed for the successful handling of both temporal segmentation and sign actions recognition tasks in \cite{PigouCGC2016} and \cite{NeverovaPAMI2015}.
And though the spatial and temporal behaviors of videos have received literature attention, there is still place for incorporating the contextual and structural information offered for instance by shape-context features \cite{ShapeContext_IPTA2015} within the deep neural architectures \cite{MutiModalSurvey2018}. 

\indent \textbf{Adaptation to challenging training conditions:}
The capability of actual CNN methods to deal with small, weakly-supervised or even previously unseen data within the learning population \cite{CGC2011Article, EPICKITCHENS2018, 2020Trend} is also an interesting challenge. 
As transfer-learning has been an attractive research topic in recent years (e.g. through SVM classifiers specialization \cite{HoudaTransferLearn_2016}), more related contributions are to be expected within future CNN/RNN architectures. Next generation contributions will have to automatically adapt their recognition schemes to the modality-types used as in \cite{NeverovaPAMI2015}, and also, to the scenes' contexts as in \cite{PfisterPose2016}.

%The good news is that not all recent contributions are hardly changing the neural architectures or their parameters. 
\indent \textbf{Actions prediction in real-time:}
Finally,  just as humans do, the optimal SL recognition solution would inevitably function in real time with the highest reliability. To do so, humans are capable of expecting the next coming behaviors by analyzing the actual attitudes of the target individuals. One promising path is the incorporation of language models as an equivalent component to the visual features extracted.  
Also, updating existing natural language processing methods to the SL grammar could help these predictions.
This dual challenge of actions prevision and on-line multi-modal streams handling has been considered by some promising contributions: 

\begin{enumerate}[i]
\item In \cite{PrevisionAction2019}, the authors predicted both the interactions labels and motion trajectories happening between pedestrians and cars. 
%within multi-person/object 
They employed CNN to segment the scene contents and describe the surroundings of the persons found, then forwarded this to LSTM networks to encode the final scene-person features. Both classification and regression layers were used for robust predictions. 
\item In \cite{PrevisionKitche20019}, Ke et al. obtained state-of-the-art actions prediction within short and long ranges on the EPIC-Kitchen dataset. They applied conditional neural connections skipping to keep only the prominent temporal features within a one-shot leaning scheme. 
\end{enumerate}

The adaptation of this technology to continuous SL recognition would enable understanding  multi-persons signing interactions. 
The creation of such a service that transcripts in real-time SL discussions is awaiting for affiliated contributions. 
% better performances could result from a same CNN architecture. 

\vspace{-0.5cm}
\subsection{Commercial solutions level}
From the previous per-modality analysis, turns out that access to the depth stream is a performance enabler for real-time solutions. The access to better sensors with miniaturized sizes and wider ranges is expected in coming years \cite{BigHand2Mdataset2017}. 
%This is for instance the case from Microsoft's Kinect to Intel's actual sensors \cite{BigHand2Mdataset2017}. 
Interestingly, as last-generation smart-phones are already equipped with IR depth sensors for facial authentication, pushing further their exploitation for FEs and SL recognition is very promising. 
With the recent mobile computation capabilities, there is also place for generating depth maps from the video's intrinsic motion as applied in \cite{DepthMotion2018, octi2018} for augmented reality experiences.
Once the milestone of depth generation achieved, multi-modal SL implementations will be unlocked. 
%adding RGB and skeletal joints will be accessible for visual-based mobile services. 

On another level, from the previously listed commercial solutions, appears the need for extending existing products (mostly focused on ASL and BSL \cite{SignAllApp2018, motionSavvy}) to cover a wider range of SL lexicons. 
It is also possible to extend the actual robotic platforms \cite{SEER_FacialRobot, ENIGMA_autismRobot} to consider SL interactions. 
The versatility of SL communities (see Tab.\ref{tab:SLdatasets}) requires adapting the actual artificial intelligence models to learn multiple SL lexicons and provide more scalable services. 

The Deaf and hearing-impaired communities have also numerous demands for dedicated interactive contents including: Educational scenarios as within the StorySign initiative \cite{stroySign2019}, communication as aimed by the in-progress SignAll solution \cite{SignAllApp2018} or even entertainment as within the OCTI mobile application \cite{octi2018}. Still, to mention a few to our best knowledge, no proper SL automatic transcription services exist for TV broadcasts, movies or outdoor facilities. 
The concerned population's scale and economic valuation  are by far bigger than the actual numeric services \cite{WorldHealthOrganization}. 
%And though initiatives exist for the creation of interactive robots capable of FEs and gestures recognition , they are still far from perfectly understanding the human natural behavior. Thus their adaption to SL interaction also awaits. 

%\vspace{-0.5cm}
%\section{Conclusion}
%\vspace{-0.5cm}
% {
% A discussion summarizing the latest accessible trends has been presented in the end.  Our literature analysis and discussion highlighted different numerical services that could be adopted both for future research works and commercial solutions. 
% 	}

%\bibpunct{(}{)}{;}{a}
\bibliographystyle{apalike}%apalike
%\bibliographystyle{spbasic}[28]
%\bibliography{general,od,stream,myproj,concept,sample1}
%\bibliography{sample1_Pham}
\bibliography{bibliographie}
\end{document}